\documentclass[conference]{IEEEtran}
\IEEEoverridecommandlockouts
\usepackage{cite}
\usepackage{amsmath,amssymb,amsfonts}
\usepackage{graphicx}
\usepackage{textcomp}
\usepackage{xcolor}
\def\BibTeX{{\rm B\kern-.05em{\sc i\kern-.025em b}\kern-.08em
    T\kern-.1667em\lower.7ex\hbox{E}\kern-.125emX}}

\usepackage{algorithm}
\usepackage{algorithmicx}
\usepackage[noend]{algpseudocode}
\usepackage[caption=false,font=normalsize,labelfont=sf,textfont=sf]{subfig}
\usepackage{amsmath}
\usepackage[absolute]{textpos}

\begin{document}

\begin{textblock}{12}(2,0.3)
	\noindent Please cite as follows: K. Malialis, M. Roveri, C. Alippi, C. G. Panayiotou, M. M. Polycarpou. A Hybrid Active-Passive Approach to Imbalanced Nonstationary Data Stream Classification. In IEEE Symposium Series on Computational Intelligence (SSCI), 2022.
\end{textblock}

\title{A Hybrid Active-Passive Approach to Imbalanced Nonstationary Data Stream Classification
\thanks{KM, CGP, MMP: This work has been supported by the European Research Council (ERC) under grant agreement No 951424 (Water-Futures), by the European Union’s Horizon 2020 research and innovation programme under grant agreements No 883484 (PathoCERT) and No 739551 (TEAMING KIOS CoE), and from the Republic of Cyprus through the Deputy Ministry of Research, Innovation and Digital Policy.}
}
\author{
	\IEEEauthorblockN{
		Kleanthis Malialis\textsuperscript{a, b},
		Manuel Roveri\textsuperscript{c},
		Cesare Alippi\textsuperscript{c, d},
		Christos G. Panayiotou\textsuperscript{a, b}
		Marios M. Polycarpou\textsuperscript{a, b}
	}\\
	\IEEEauthorblockA{
		\textsuperscript{a} \textit{KIOS Research and Innovation Center of Excellence}\\
		\textsuperscript{b} \textit{Department of Electrical and Computer Engineering}\\
		\textit{University of Cyprus}, Nicosia, Cyprus\\
		Email: \{malialis.kleanthis, christosp, mpolycar\}@ucy.ac.cy\\
		ORCID: \{0000-0003-3432-7434, 0000-0002-6476-9025, 0000-0001-6495-9171\}
	}\\

	\IEEEauthorblockA{
	\textsuperscript{c} \textit{Dipartimento di Elettronica, Informazione e Bioingegneria}\\
	\textit{Politecnico di Milano}, Milan, Italy\\
	Email: \{manuel.roveri, cesare.alippi\}@polimi.it\\
	ORCID: \{0000-0001-7828-7687, 0000-0003-3819-0025\}
	}\\
	
	\IEEEauthorblockA{
	\textsuperscript{d} \textit{Università della Svizzera italiana}, Lugano, Switzerland\\
	}

}

\maketitle

\begin{abstract}
In real-world applications, the process generating the data might suffer from nonstationary effects (e.g., due to seasonality, faults affecting sensors or actuators, and changes in the users' behaviour). These changes, often called concept drift, might induce severe (potentially catastrophic) impacts on trained learning models that become obsolete over time, and inadequate to solve the task at hand. Learning in presence of concept drift aims at designing machine and deep learning models that are able to track and adapt to concept drift. Typically, techniques to handle concept drift are either active or passive, and traditionally, these have been considered to be mutually exclusive. Active techniques use an explicit drift detection mechanism, and re-train the learning algorithm when concept drift is detected. Passive techniques use an implicit method to deal with drift, and continually update the model using incremental learning. Differently from what present in the literature, we propose a hybrid alternative which merges the two approaches, hence, leveraging on their advantages. The proposed method called Hybrid-Adaptive REBAlancing (HAREBA) significantly outperforms strong baselines and state-of-the-art methods in terms of learning quality and speed; we experiment how it is effective under severe class imbalance levels too.
\end{abstract}


\section{Introduction}\label{sec:intro}
Recent advances in hardware and sensor technology, and the growth of inter-connected Internet-of-Things (IoT) devices have generated a plethora of data stream-based applications, such as, monitoring of critical infrastructure systems, smart home and cities and industry 4.0 \cite{kyriakides2014intelligent}. Learning online, i.e., as new information is becoming available under streaming conditions poses significant challenges, which hinder the practicality and deployability of learning algorithms.

A key challenge encountered is the nonstationary nature of data streams which evolve over time \cite{ditzler2015learning}. Typically, this is caused by concept drift, which refers to a change in the data distribution \cite{lu2018learning}. For example, in monitoring systems nonstationarity can be caused by hardware failures, and in security systems by changes in users' behaviour.

Traditionally, methods to address concept drift are either active or passive \cite{ditzler2015learning}. The former use an explicit drift detection mechanism (e.g., threshold-based), and re-train the learning model only when drift is detected. The latter use an implicit method to deal with drift (e.g., memory-based), and continually update the model incrementally. 
These two approaches have been been rarely integrated with very few attempts (e.g., \cite{alippi2017learning}) focusing only on specific aspects of the learning phase. 

In addition, class imbalance, which refers to the problem of having a skewed data distribution often affects data-streams \cite{wang2018systematic}. This renders a learning model incapable of performing well, particularly, of accurately predicting examples from minority classes, unless particular mitigation approaches are introduced to rebalance the learning problem. 

The goal of this work is to introduce a hybrid active-passive approach for class imbalanced data streams. In more detail, the novel contributions of this work are:
\begin{itemize}
	\item We propose a hybrid active-passive approach to handle nonstationarity and concept drift in data streams. Contrary to the traditional framework where active and passive approaches have been applied in isolation, we propose a method which combines the two in synergy to take advantage of both.

	\item We propose the Hybrid-Adaptive REBAlancing (HAREBA) method, which significantly extends AREBA \cite{malialis2021online}. The original method is passive, i.e., it implicitly handles drift by using memory and incremental learning; it also uses a dynamic mechanism to constantly maintain balance within the memory. This work synergistically incorporates an explicit drift detection mechanism to form an effective hybrid method, whose performance (in terms of classification ability) is superior to baselines and state-of-the-art methods; it significantly outperforms them in terms of learning quality and speed too.
\end{itemize}

The paper is structured as follows. Preliminary material is introduced in Section~\ref{sec:background}. Related work is presented in Section~\ref{sec:related}. HAREBA is described in Section~\ref{sec:method}. The experimental setup and results are presented in Sections~\ref{sec:exp_setup} and \ref{sec:exp_results} respectively. We conclude in Section~\ref{sec:conclusion}.

\section{Preliminaries}\label{sec:background}
\textbf{Online} learning considers a data generating process that provides at each time $t$ a sequence of examples $S = \{S^t\}_{t=1}^T$, where $S^t = \{(x^t_i,y^t_i)\}^M_{i=1}$. The number of steps is denoted by $T \in [1, \infty)$ where the data are typically sampled from a long, potentially infinite, sequence. The number of examples at each step is denoted by $M$. If $M=1$, it is termed \textbf{one-by-one online} learning, otherwise it is termed \textbf{batch-by-batch online} learning \cite{ditzler2015learning}. The examples are drawn from an unknown probability distribution $p^{t}(x,y)$, where $x^t \in \mathbb{R}^d$ is a $d$-dimensional vector in the input space $X \subset \mathbb{R}^d$, and $y^t \in \{0, 1\}$ is the class label. The focus of this paper is on binary classification and, as a convention, the positive class represents the minority class.

We focus on one-by-one learning, i.e., $S^t = (x^t, y^t)$, which is important for real-time monitoring. One-by-one learning requires the model to adapt immediately upon seeing a new example, and algorithms intended for batch-by-batch learning are, typically, not applicable for such tasks \cite{wang2018systematic}. A one-by-one online classifier receives a new instance $x^t$ at time $t$ and makes a prediction $\hat{y}^t$ based on a concept $f: X \to Y$, such that, $\hat{y}^t = f^{t-1}(x^t)$. In \textbf{online supervised} learning, the classifier receives the true label $y^t$, its performance is evaluated using a cost function $J^t$ and is then trained based on the cost incurred. This is repeated at each step. The gradual adaptation of the classifier without complete re-training $f^t = f^{t-1}.train(J^t)$ is termed \textbf{incremental} learning \cite{losing2018incremental}.

Since data are sampled from a long, potentially infinite, sequence which is typically the case in data streams, it is unrealistic to expect that all acquired labels will be available at all times. The classifier should use a fixed amount of memory for data storage. If learning occurs on the most recent example, without using a memory, it is termed \textbf{one-pass} learning \cite{ditzler2015learning}. In such case, the cost at time $t$ is calculated using the loss function $l$ as $J^t=l(y^t,f^{t-1}(x^t))$.

Algorithms belonging to this learning paradigm are typically trained by supervision, i.e., from user interaction by domain experts. Various and widely studied domains exist that fit into this paradigm, e.g., in finance, security, recommender systems, and environmental monitoring \cite{ditzler2015learning, wang2018systematic}. This paradigm may not be ideal in some cases; we direct the interested reader to alternatives ones, such as, online active learning (e.g. \cite{zliobaite2013active, malialis2022nonstationary, malialis2022augmented, malialis2020data}) as this is outside our scope.

A key challenge encountered in some applications is that of data \textbf{nonstationarity} \cite{ditzler2015learning, lu2018learning}, typically caused by \textbf{concept drift}, which represents a change in the joint probability. The drift between steps $t_i$ and $t_j$, where $j > i$, is defined as:
\begin{equation}
\quad p^{t_i}(x,y) \neq p^{t_j}(x,y)
\end{equation}

Abrupt changes in the probability density function implies a sudden variation from $p^{t_i}$ to $p^{t_j}$, while gradual drift refers to a smooth transition from $p^{t_i}$ to $p^{t_j}$.


\section{Related Work}\label{sec:related}

\subsection{Concept drift adaptation}\label{sec:related_drift}

\subsubsection{\textbf{Passive} methods}
These methods implicitly address the problem and include memory-based and ensembling methods. Memory-based methods, typically, use a sliding window to store the most recent examples, which a classifier is trained on. A representative method and one of the earliest is FLORA \cite{widmer1996learning}. A key challenge is to determine a priori the window size; to address this, methods use an adaptive sliding window \cite{widmer1996learning} or multiple windows \cite{lazarescu2004using}. Ensembling refers to a dynamic set of classifiers, in which classifiers can be added or discarded based on their performance. Representative methods are Learn++.NSE \cite{elwell2011incremental}, Accuracy Updated Ensemble (AUE) \cite{brzezinski2013reacting}, and Diversity for Dealing with Drifts (DDD) \cite{minku2011ddd}.

\subsubsection{\textbf{Active} methods}
These methods are also referred to as change detection-based, and use explicit mechanisms to detect concept drift, such as, statistical tests (e.g., Just-In-Time -JIT- classifiers \cite{alippi2008justI, alippi2008justII, alippi2013just}), and threshold-based mechanisms (e.g., Drift Detection Method (DDM) \cite{gama2004learning} and Early DDM (EDDM) \cite{baena2006early}). Threshold-based methods monitor the number of errors made after each prediction. When particular thresholds are surpassed, the system raises Warning and Drift alarms.

For a comprehensive survey of concept drift methods we direct the interested reader towards these excellent papers \cite{ditzler2015learning, gama2014survey, lu2018learning}. For a survey with a focus on ensembling the interested reader is directed towards \cite{krawczyk2017ensemble, gomes2017survey}.

Typically, passive methods use incremental learning. Unlike passive methods, active methods don't continually update the classifier, but instead perform a complete re-training every time drift is detected. In this work, we use a threshold-based method which works in close synergy with incremental learning to achieve a hybrid active-passive collaboration.

\subsection{Class imbalance}\label{sec:related_imbalance}
The previously described methods do not take into account the imbalance problem. These should be combined with class imbalance methods to address the combined problem \cite{wang2018systematic}.

\subsubsection{\textbf{Cost-sensitive learning}}
These methods alter the loss function to induce different penalties for example misclassifications of different classes. The \textit{Cost-Sensitive Online Gradient Descent (CSOGD)} algorithm uses the loss function:
\begin{equation}\label{eq:cs_cost}
J = ( I_{y^t=0} + I_{y^t=1} \frac{\gamma_p}{\gamma_n} ) ~ l(y^t, \hat{y}^t),
\end{equation}
\noindent where $I_{condition}$ is the identify function that returns 1 if $condition$ is satisfied and 0 otherwise, $\gamma_p, \gamma_n \in [0,1]$ and $\gamma_p + \gamma_n = 1$ are the misclassification costs for positive and negative classes respectively \cite{wang2014cost}. A weakness is that the costs must be pre-determined, however, the imbalance level is not known a-priori. One way to calculate the imbalance level on-the-fly is the Class Imbalance Detection (CID) \cite{wang2013learning}, which calculates a time-decayed metric for each of the two classes.


\subsubsection{\textbf{Resampling}}
These methods alter the training set (e.g., via oversampling and undersampling) to deal with the skewed data distribution \cite{wang2018systematic}.

Oversampling-based Online Bagging (OOB) \cite{wang2015resampling} is an ensemble method that uses the CID method (described earlier) to determine the minority and majority classes on-the-fly. It adjusts the learning bias from the majority to the minority class adaptively through resampling by updating each classifier k times, specifically, when a minority class example is observed the value of k increases, otherwise its value is decreased. OOB uses incremental learning, and it is a one-pass learner.

Adaptive REBAlancing (AREBA) \cite{malialis2021online, malialis2018queue} effectively addresses the dual-problem of concept drift and class imbalance. It uses a passive approach (memory-based) to handle drift, and a dynamic rebalancing mechanism to handle imbalance. As this work proposes a significant extension to AREBA, we will briefly describe the original method in Section~\ref{sec:proposed_passive}.

Other methods, include, SElective Recursive Approach (SERA) \cite{chen2009sera}, and Learn++CDS and Learn++NIE \cite{ditzler2013incremental}.

\section{HAREBA}\label{sec:method}
HAREBA or Hybrid-AREBA uses incremental learning and a threshold-based explicit drift detection. It continually monitors the prediction errors (i.e., the errors in classifying the input samples $x^t$), and raises \textit{warning} and \textit{drift alarms} when the average prediction error exceeds certain thresholds. These threshold-based detection mechanisms will be detailed in what follows. Recall that in our framework, the algorithm at each step $t$ observes an example $x^t$ and makes a prediction $\hat{y}^t \in [0,1]$ and the class label is then acquired $y^t \in [0,1]$, as shown in Lines 11 - 14 of Algorithm~\ref{alg:hybrid-areba}. Then Lines 15 - 36 and Lines 37 - 45 introduce the active and passive component of HAREBA, respectively. The proposed algorithm is detailed and commented in what follows.

\subsection{Active approach}
\textbf{Error monitoring}. The error score is calculated by $score^t = I_{y^t == \hat{y}^t}$, where $I$ is the identify function; it is also referred to as the 0/1 score. A queue $q^t_{scores}$ is maintained of size $W_{scores}$ to store the recent prediction scores. This corresponds to Lines 16 - 17 in Algorithm~\ref{alg:hybrid-areba}.

\textbf{Error modelling}. We model the errors using a Binomial distribution as in \cite{gama2004learning} with estimated parameter $p^t = \frac{1}{|q^t_{scores}|} \sum_{s \in q^t_{scores}} s$. The mean and standard deviation are:

\begin{equation}
\begin{aligned}
&	\mu^t = p^t\\
&	\sigma^t = \sqrt{\frac{p^t (1 - p^t)}{|q^t_{scores}|}}
\end{aligned}
\end{equation}

\textbf{Thresholds}. Given pre-specified values $\beta_{improve}, \beta_{warn}, \beta_{drift}$ we calculate the following thresholds:
\begin{equation}\label{eq:thresholds}
\begin{aligned}
	\theta^t_{improve} &= \mu^t + \sigma^t * \beta_{improve}\\
	\theta^t_{warn} &= \mu^t - \sigma^t * \beta_{warn}\\
	\theta^t_{drift} &= \mu^t - \sigma^t * \beta_{drift}
\end{aligned}
\end{equation}

The thresholds should satisfy $\beta_{warn} < \beta_{drift}$, while typically $\beta_{improve} \sim= \beta_{warn}$; for example, $\beta_{drift} = 5.0$, $\beta_{warn} = 3.0$, and $\beta_{improve} = 2.5$. The thresholds are fist calculated in Line 19 - 20. Surpassing these will trigger various events which are shown in blue font in Algorithm~\ref{alg:hybrid-areba} and are described below. The thresholds are task-specific.

\begin{algorithm}[t!]
	\caption{HAREBA}
	\label{alg:hybrid-areba}
	\begin{algorithmic}[1]
		
		\Statex \textbf{Arguments:}
		\State $B \geq 2$ \Comment Total memory for AREBA
		\State $W_{scores}$ \Comment{Window size of $q_{scores}$}
		\State $W_{dd}$ \Comment{Window size of $q_{dd}$}
		\State $waiting\_time$ \Comment Until Active part starts
		\State $expire\_time$ \Comment For the Warning flag
		
		\Statex \textbf{Start:}
		\State $f^0.init()$ \Comment Initialise NN
		\State $q^0_p, q^0_n = \{\}$ \Comment AREBA queues of max. capacity $\frac{B}{2}$
		\State $q_{scores} = \{\}$ \Comment Queue of size $W$ with 0/1 scores
		\State $flag_{warning} = False$ \Comment Warning flag
		\State $q_{dd} = \{\}$ \Comment Queue of examples after $flag_{warning}$ raised
		
		\For{each time step $t \in [1,\infty)$}
		
		\State{receive example $x^t \in \mathbb{R}^d$}
		\State{predict class $\hat{y}^t \in \{0,1\}$}
		\State{receive true label $y^t \in \{0,1\}$}
		
		\If{$t \geq waiting\_time$} \Comment \textcolor{red}{Active approach}
		
		\State $score^t = I_{y^t == \hat{y}^t}$ \Comment Get 0/1 loss score
		\State $q_{scores}^t = q_{scores}^{t-1}.append(score^t)$
		
		\If{$q^t_{scores}.isFull()$}
		
		\If{$q^t_{scores}$ full for the first time \textbf{or} here after a Drift flag reset}
		\State set $\theta_{improve}, \theta_{warn}, \theta_{drift}$ \Comment Eq. (\ref{eq:thresholds})
		\EndIf
		
		\If{$\mu^t \geq \theta_{improve}$} \Comment \textcolor{blue}{Model improved}
		\State re-set $\theta_{improve}, \theta_{warn}, \theta_{drift}$ \Comment Eq. (\ref{eq:thresholds})
		\State $flag_{warn} = False$
		\EndIf
		
		\If{$\mu^t \leq \theta_{warn}$ \textbf{or} $flag_{warn}$} \Comment \textcolor{blue}{Warning}
		\State $q^t_{dd} = q^{t-1}_{dd}.append(x^t)$
		\If{expiration time passed} \Comment reset flag
		\State $q^t_{dd} = \{\}$
		\State $flag_{warn} = False$
		\EndIf
		\EndIf
		
		\If{$\mu^t \geq \theta_{drift}$} \Comment \textcolor{blue}{Drift}
		\State $f^{t-1} = f^{t-1}.init()$ \Comment new classifier
		\State calculate cost $C^t$ on $q_{dd}^t$
		\State $f^t = f^{t-1}.train(C^t)$ \Comment train classifier
		\State $q^t_p, q^t_n = \{\}$ \Comment reset AREBA queues
		\State $q^t_{scores} = \{\}$
		\State $q^t_{dd} = \{\}$
		\State $flag_{warn} = False$ \Comment reset Warning flag
		\EndIf
		
		\EndIf
		
		\EndIf
		
		\If{$t < waiting\_time$ \textbf{or} $flag_{warn} == False$} \Comment \textcolor{red}{Passive approach}
		
		\If{$y^t == 1$}
		\State{$q^t_p = q^{t-1}_p.append\big((x^t,y^t)\big)$}
		\Else
		\State{$q^t_n = q^{t-1}_n.append\big((x^t,y^t)\big)$}
		\EndIf
		
		\State adjust capacities of $q^t_p$, $q^t_n$ \Comment AREBA mechanism
		\State prepare the training set $q^t = q^t_p \cup q^t_n$
		\State calculate cost $J^t$ on $q^t$
		\State update classifier once $f^t = f^{t-1}.train(J^t)$
		
		\EndIf
		
		\EndFor
		
	\end{algorithmic}
\end{algorithm}

\textbf{Model improvement}.
 This stage (Lines 21 - 23) indicates that the model's performance has improved. This can happen due to the passive nature of the method, i.e., incremental learning (described later). It occurs when the following condition is satisfied:
 \begin{equation}
	\mu^t \geq \theta^t_{improve}
\end{equation}
 
The thresholds are re-calculated using Eq. (\ref{eq:thresholds}), and the Warning flag is reset (if raised), as shown in Lines 21 - 23.

\textbf{Warning flag}. This stage (Lines 24 - 28) raises an alarm of a possible concept drift when the following condition is satisfied:
\begin{equation}
\mu^t \leq \theta^t_{warn}
\end{equation}

From this time forward, we start storing the most recently arriving example $x^t$ and subsequent ones to the queue $q^t_{dd}$ of size $W_{drift}$. If the Warning flag is raised for a long time without a Drift alarm being raised, then it is reset, as well as $q^t_{dd}$. This corresponds to Lines 24 - 28.

\textbf{Drift alarm}. This stage (Lines 29 - 36) triggers a concept drift alarm when the following condition is satisfied:
\begin{equation}
\mu^t \leq \theta^t_{drift}
\end{equation}

At this point, the current classifier is discarded a new one is initiated $f^{t-1} = f^{t-1}.init()$ (Line 30). We calculate the cost $C^t$ (Line 31) incurred on the examples in $q^t_{dd}$ which is defined as follows:
\begin{equation}
    C^t = \frac{1}{|q^t_{dd}|} \sum_{(x_i,y_i) \in q^t_{dd}} l(y_i, f^{t-1}(x_i))
\end{equation}
The new classifier is trained based on the loss incurred $f^t = f^{t-1}.train(C^t)$ (Line 32). Moreover, the AREBA mechanism is reset, i.e., the two AREBA queues are reset (Line 33). Also, the Warning flag is reset, as well as the queues $q^t_{scores}$ and $q^t_{dd}$ (Lines 34 - 36).

\subsection{Passive approach}\label{sec:proposed_passive}
The algorithm is in Passive mode when $flag_{warn}$ is not raised (Line 37).

\textbf{Memory-based}. It selectively includes in the training set a subset of the minority (positive) and majority (negative) examples by storing them in the queues $q^t_n$ and $q^t_p$ respectively (Lines 38 - 41). Due to the memory-based nature, it constitutes an implicit (passive) way to address concept drift.

\textbf{Rebalancing}. Furthermore, it incorporates a dynamic mechanism to adaptively modify the queue sizes to maintain class balance within the queues. Specifically, AREBA dynamically resizes the two queues $q^t_n$ and $q^t_p$ according to the most recent class label. Without this mechanism, the initial class imbalance problem would still persist in the memory (queue)-based system. This is stated in Line 42 and the details are provided in the original paper \cite{malialis2021online}.

\textbf{Incremental learning}. The cost function $J^t$ is calculated based on the examples in $q^t_n$ and $q^t_p$. Incremental learning is then used, that is, the gradual adaptation of the classifier without complete re-training is performed $f^t = f^{t-1}.train(J^t)$ as shown in (Lines 43 - 45).

 \begin{figure}[t]
 	\centering
 	
 	\subfloat[circle]{\includegraphics[scale=0.18]{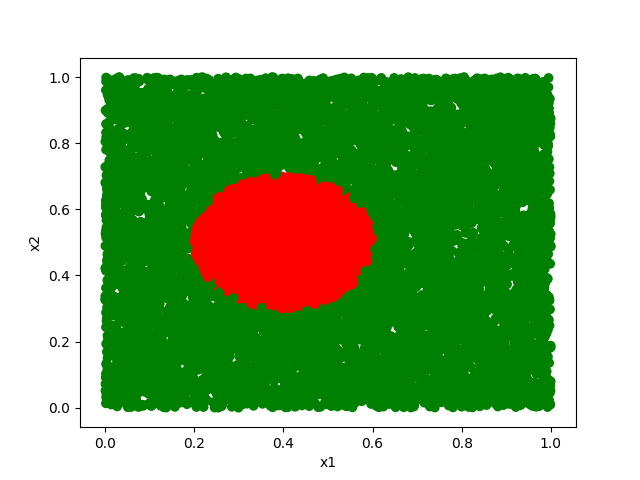}%
 		\label{fig:circle}}
 	\subfloat[sine]{\includegraphics[scale=0.18]{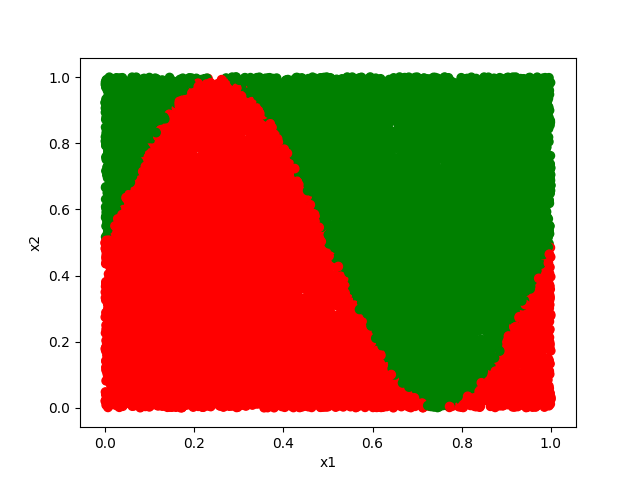}%
 		\label{fig:sine}}
 	\subfloat[sea]{\includegraphics[scale=0.18]{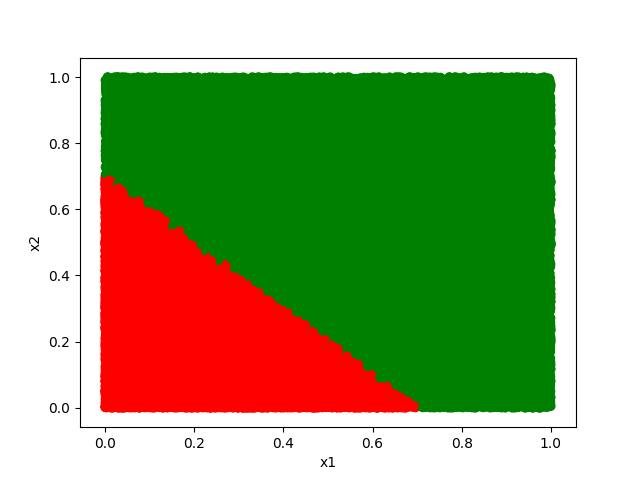}%
 		\label{fig:sea}}
 	
 	\caption{The three synthetic datasets used in our study.}
 \end{figure}
 
\section{Experimental Setup}\label{sec:exp_setup}

\begin{figure*}[t]
	\centering
	
	\subfloat[Baseline]{\includegraphics[scale=0.11]{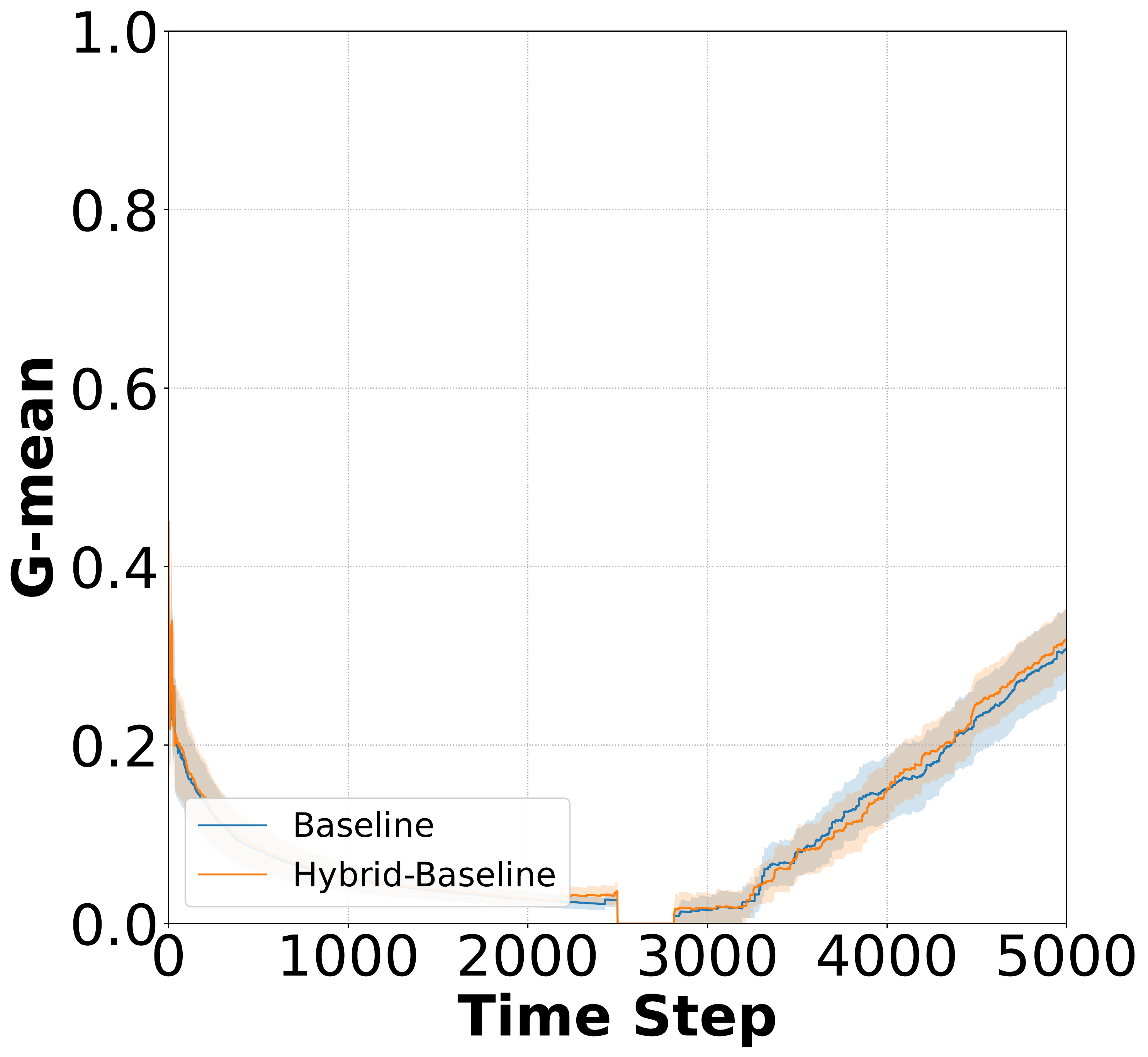}%
		\label{fig:circle01_drift_baseline}}
	\subfloat[Sliding]{\includegraphics[scale=0.11]{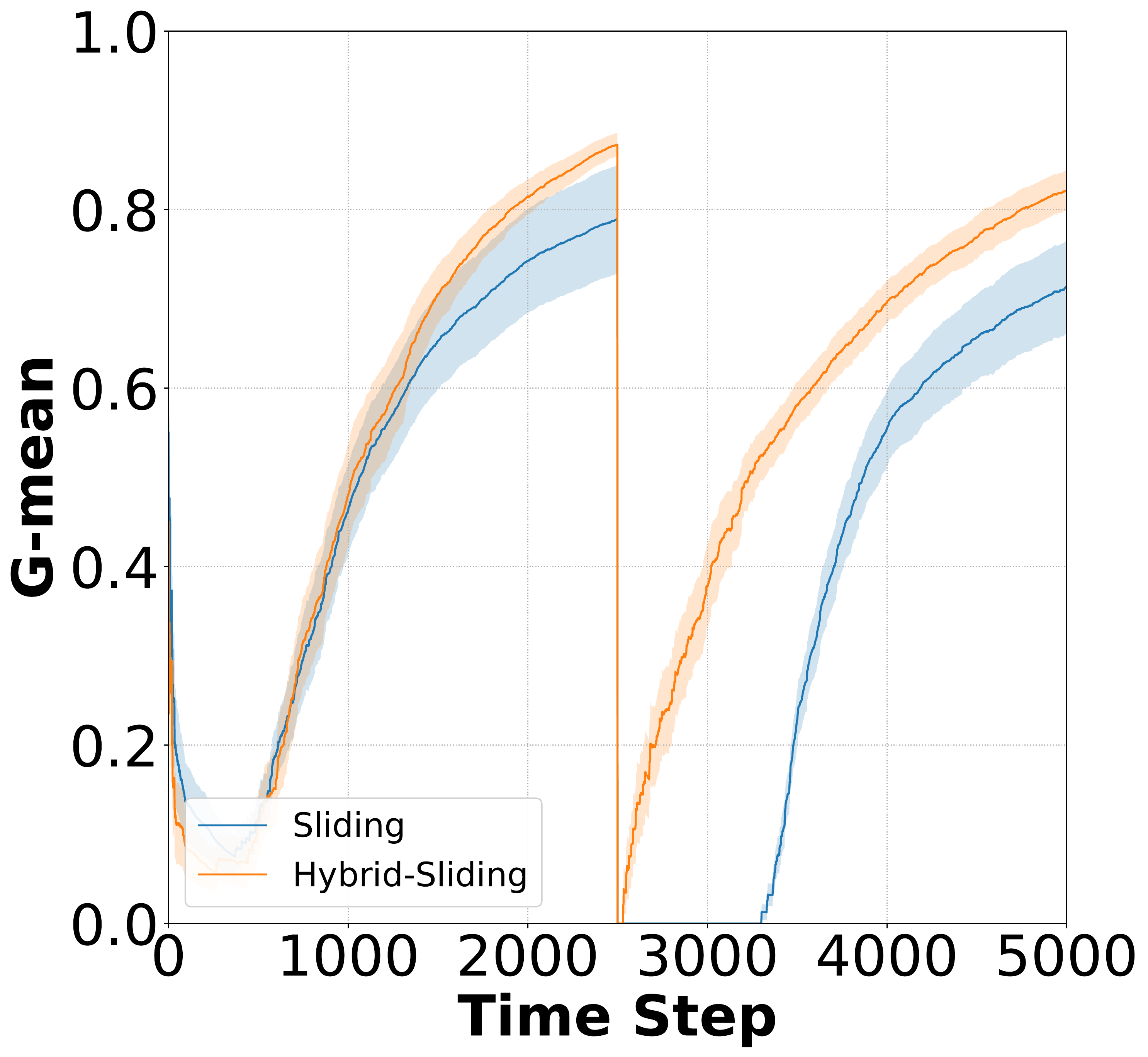}%
		\label{fig:circle01_drift_sliding}}
	\subfloat[Adaptive\_CS]{\includegraphics[scale=0.11]{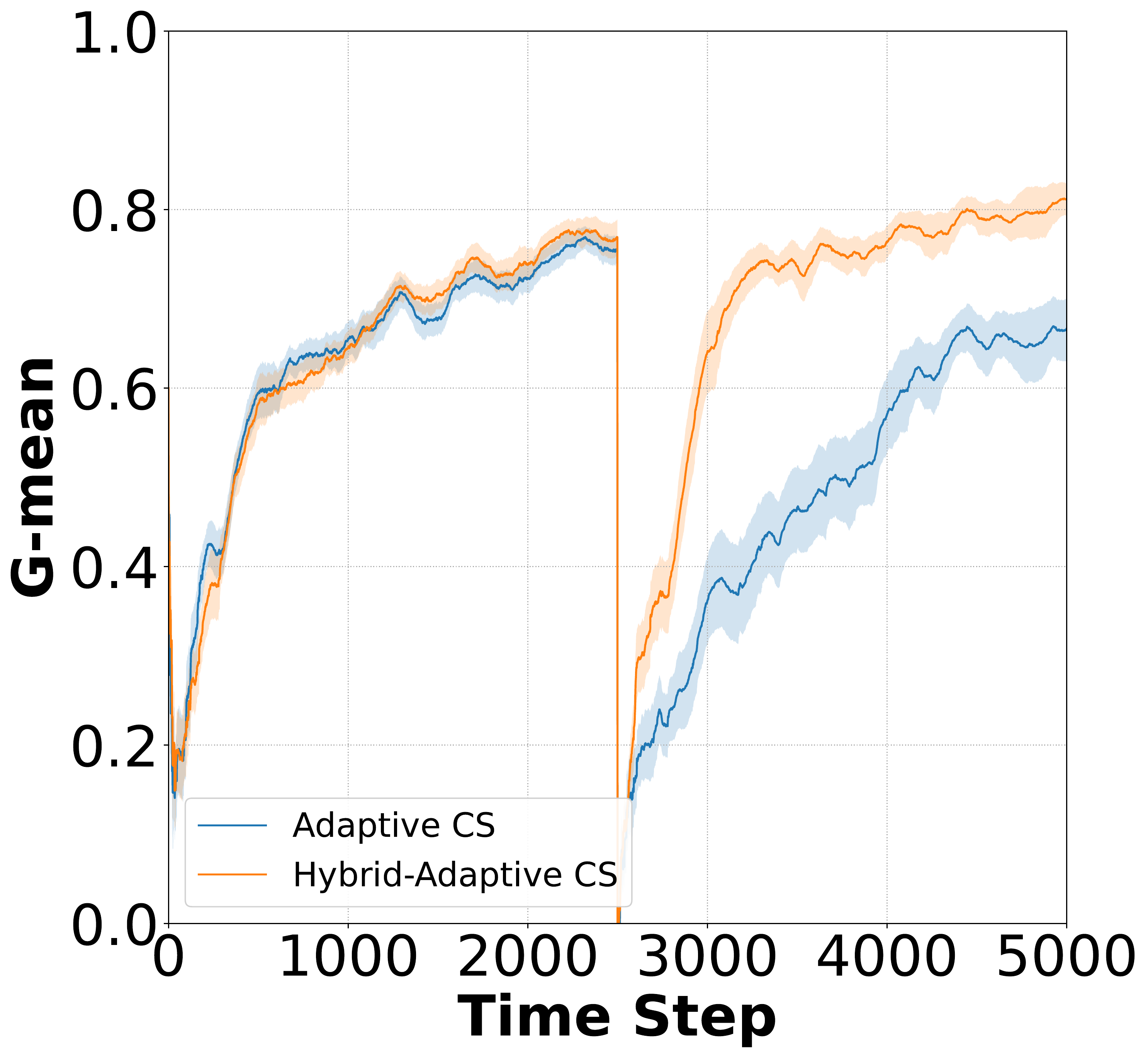}%
		\label{fig:circle01_drift_adaptive_cs}}
	\subfloat[OOB]{\includegraphics[scale=0.11]{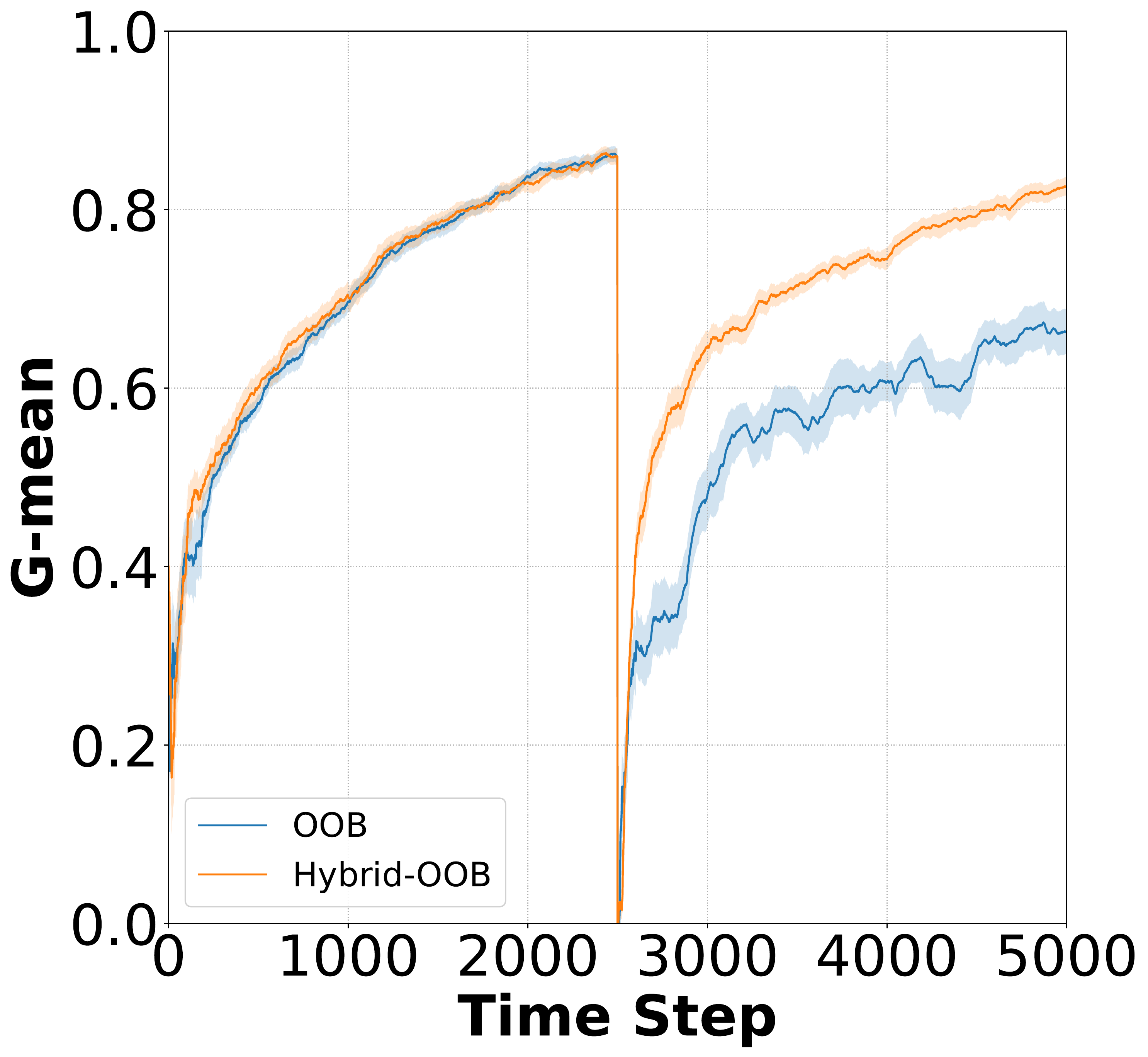}%
		\label{fig:circle01_drift_oob}}
	\subfloat[AREBA]{\includegraphics[scale=0.11]{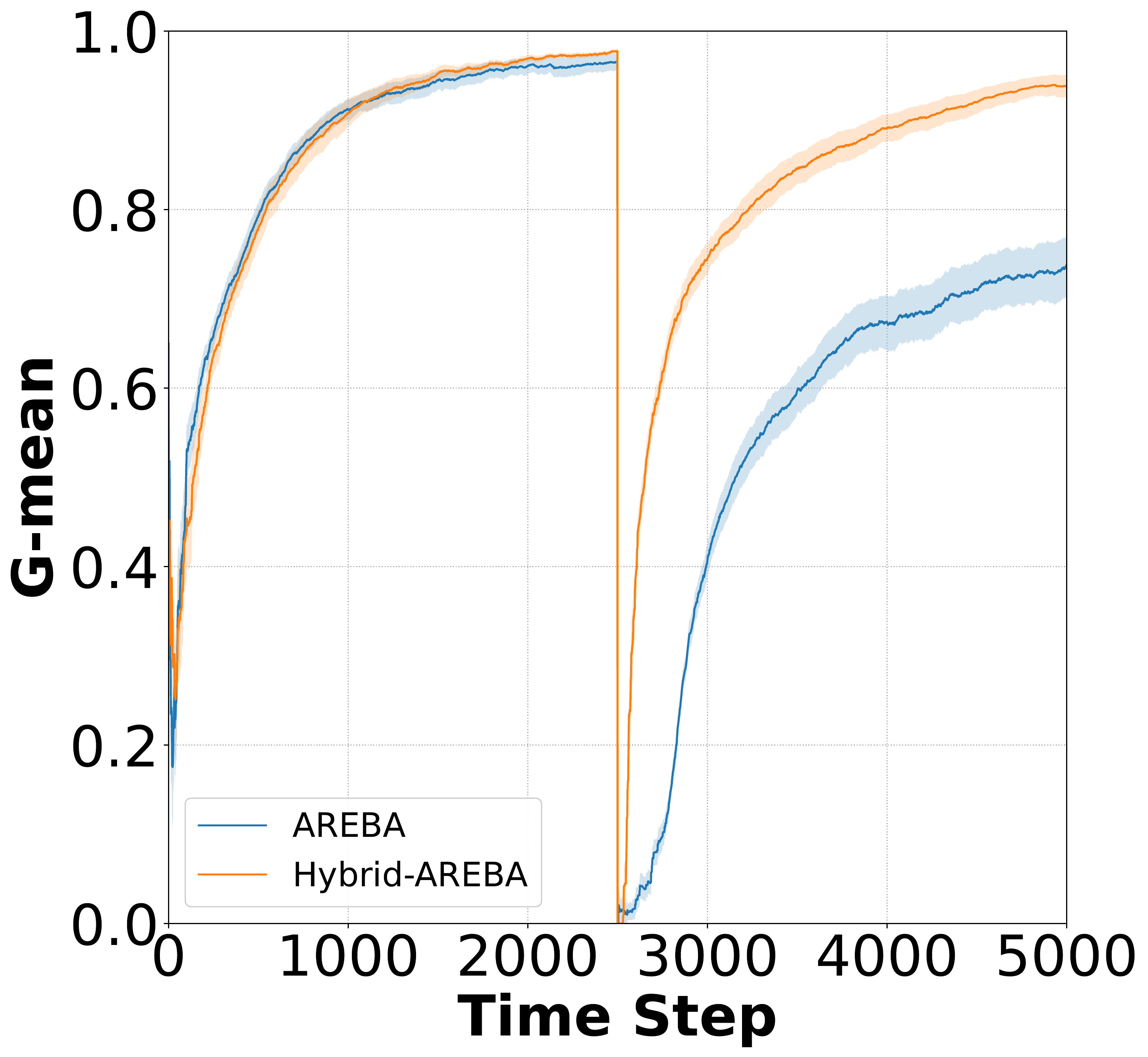}%
		\label{fig:circle01_drift_areba}}
	
	\caption{The role of hybrid learning in the Circle dataset}\label{fig:role_circle}
\end{figure*}

\begin{figure*}[t]
	\centering
	
	\subfloat[Baseline]{\includegraphics[scale=0.11]{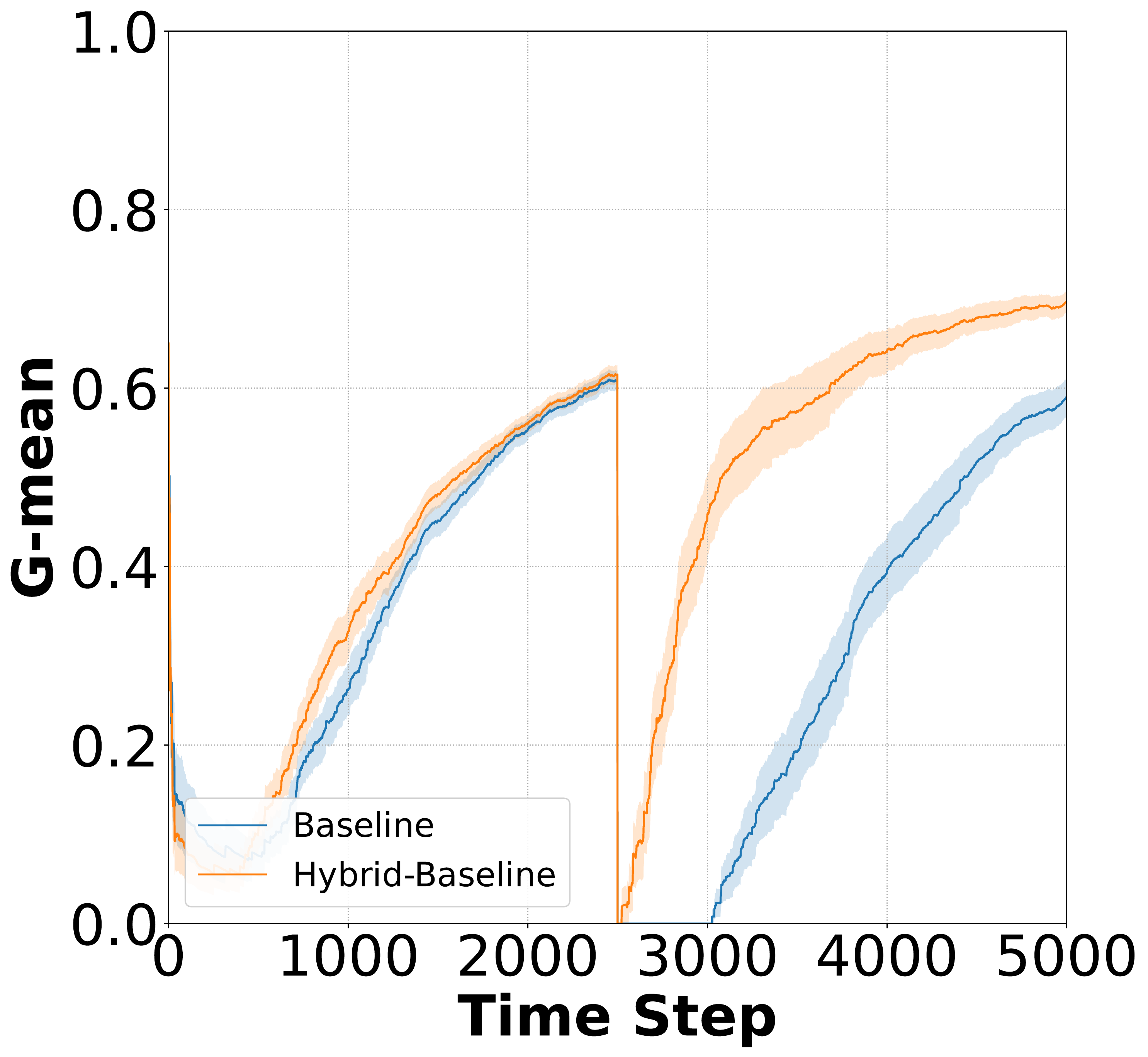}%
		\label{fig:sine01_drift_baseline}}
	\subfloat[Sliding]{\includegraphics[scale=0.11]{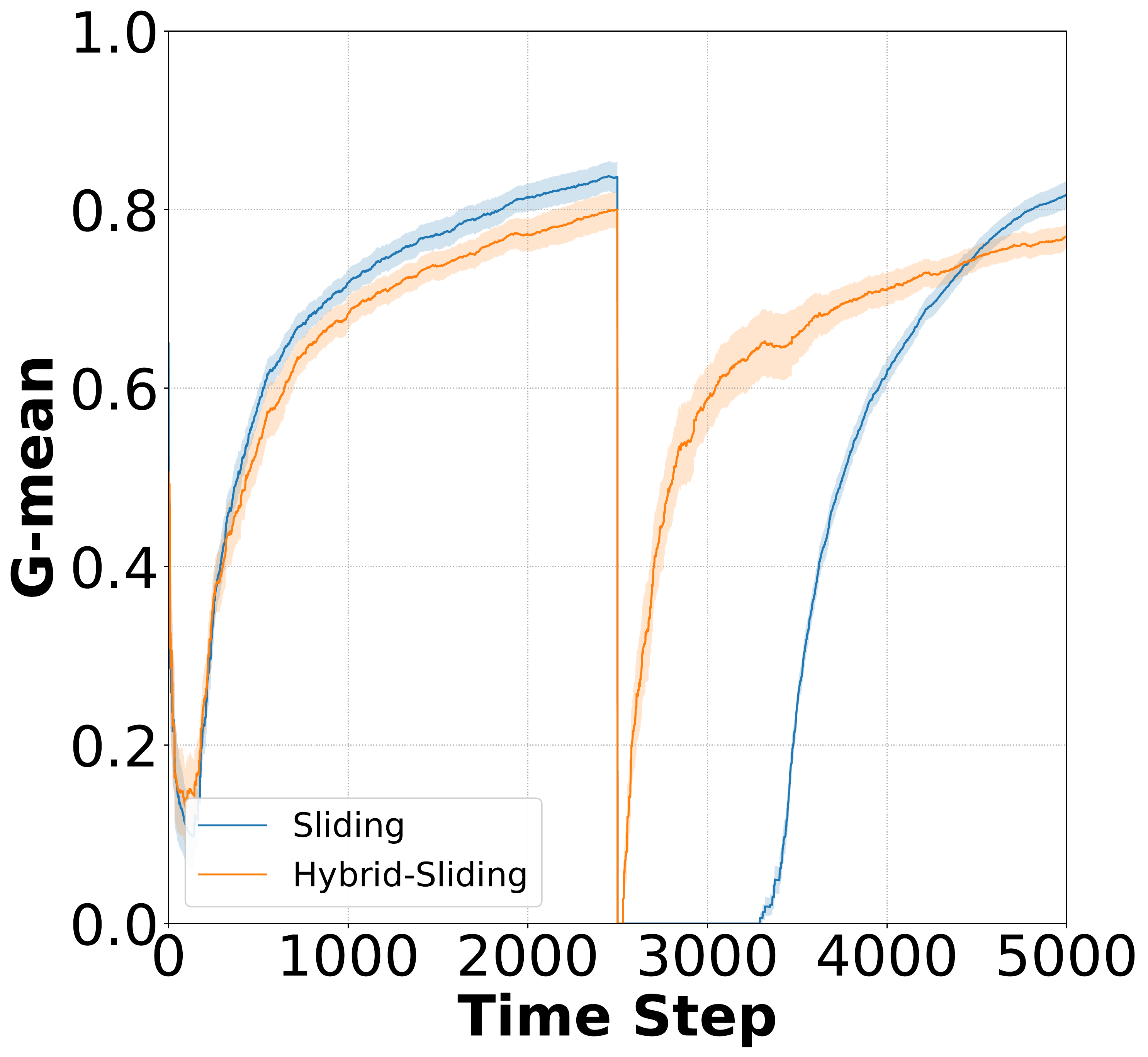}%
		\label{fig:sine01_drift_sliding}}
	\subfloat[Adaptive\_CS]{\includegraphics[scale=0.11]{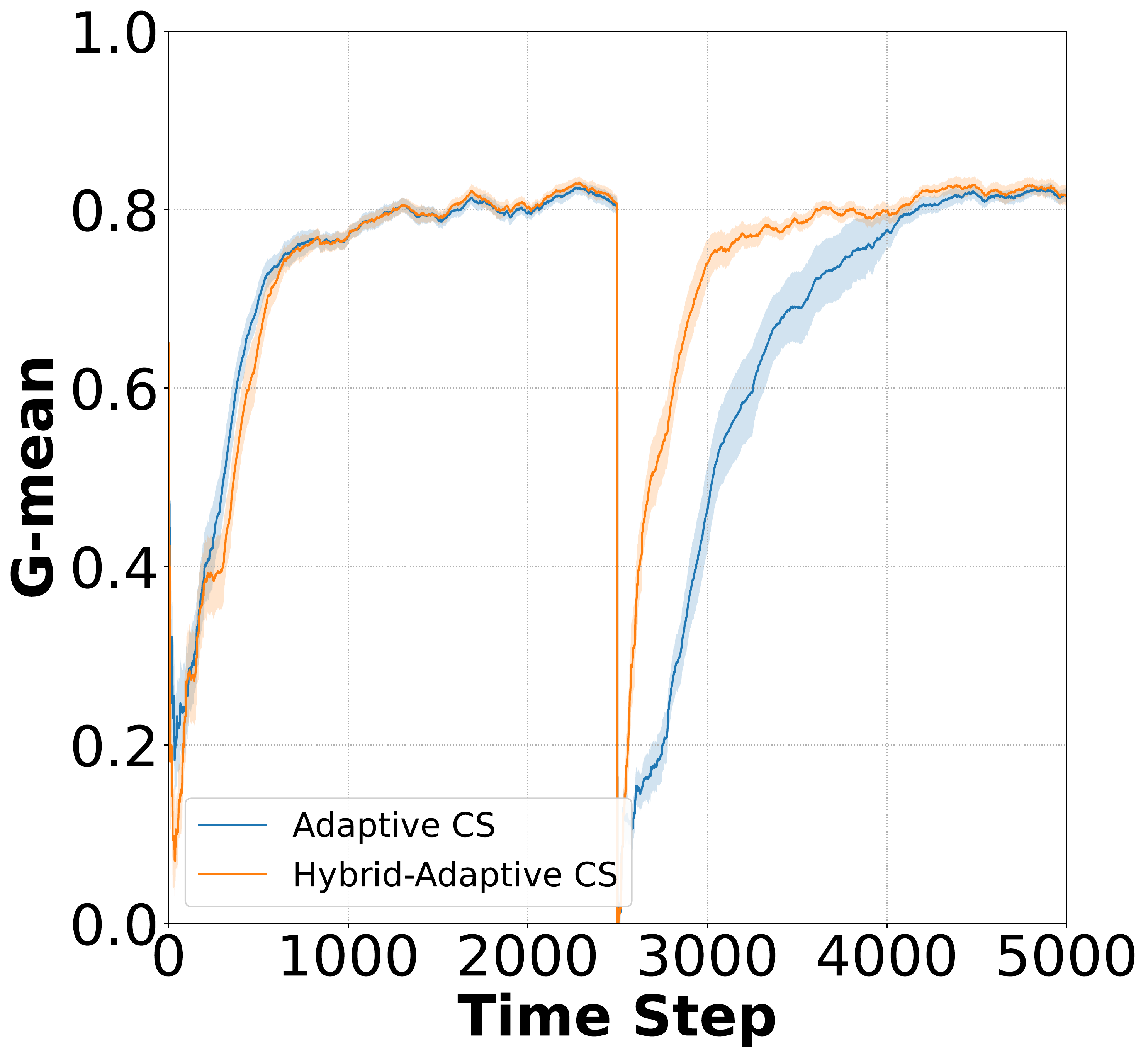}%
		\label{fig:sine01_drift_adaptive_cs}}
	\subfloat[OOB]{\includegraphics[scale=0.11]{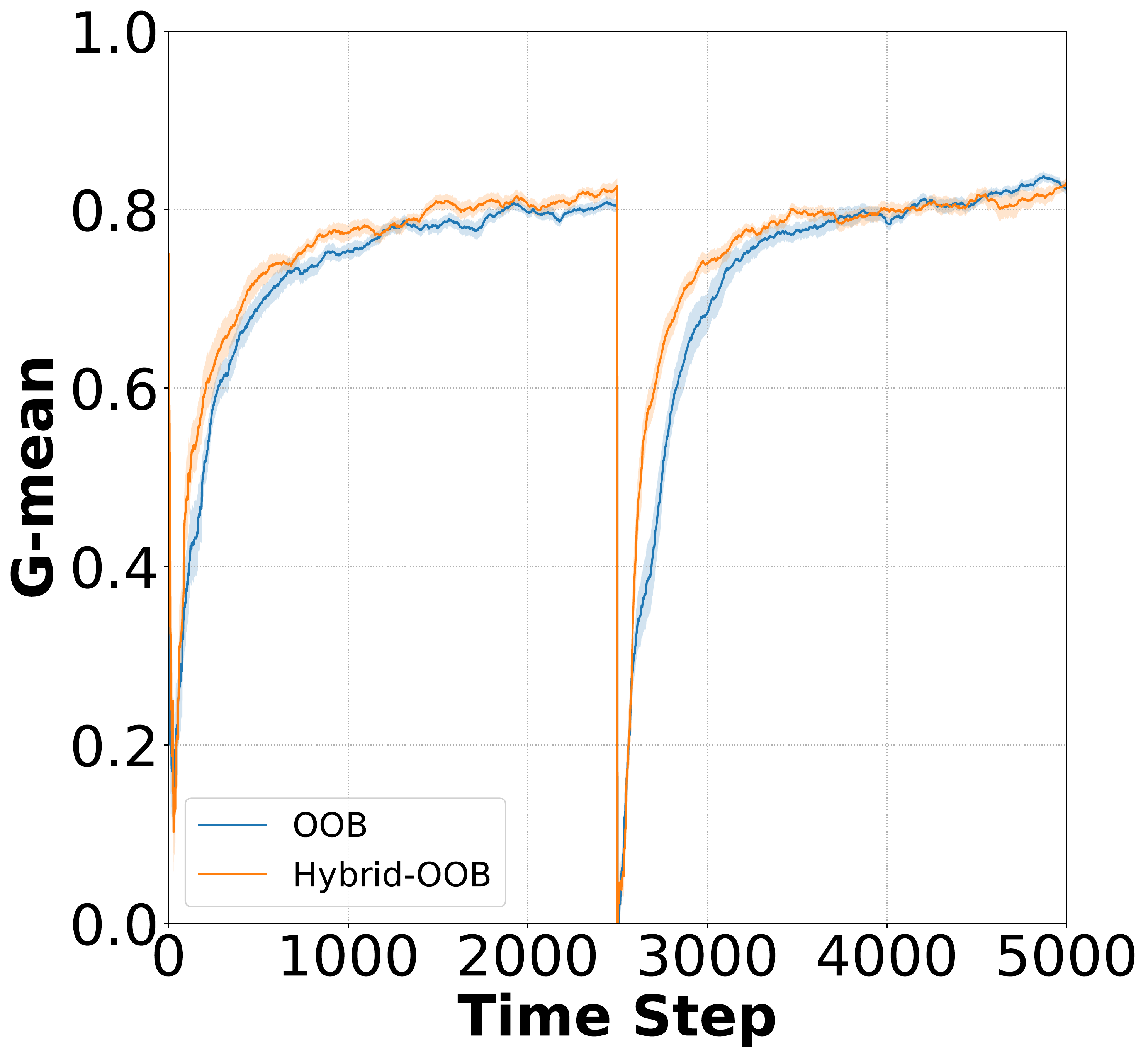}%
		\label{fig:sine01_drift_oob}}
	\subfloat[AREBA]{\includegraphics[scale=0.11]{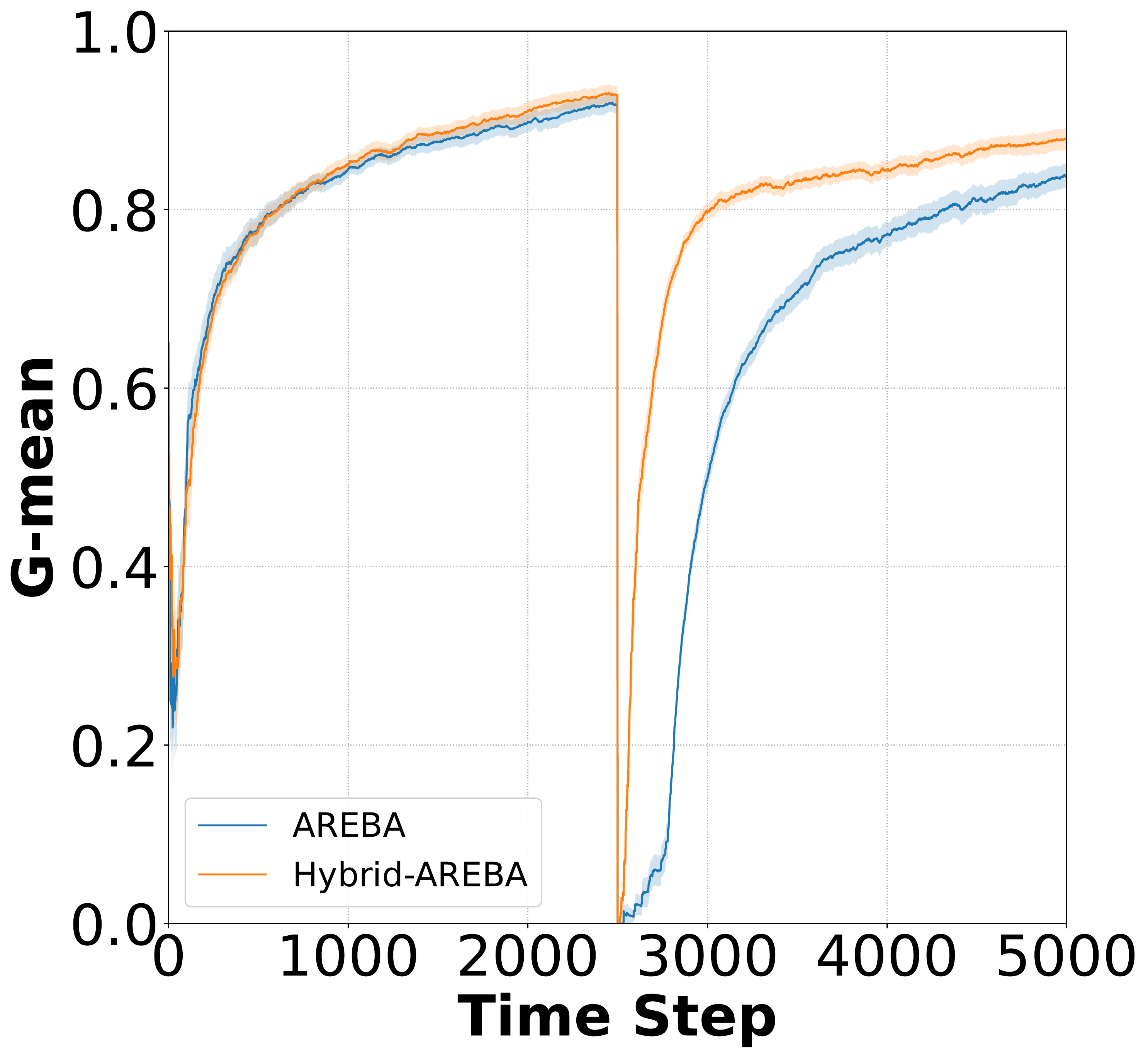}%
		\label{fig:sine01_drift_areba}}
	
	\caption{The role of hybrid learning in the Sine dataset}\label{fig:role_sine}
\end{figure*}

\begin{figure*}[t]
	\centering
	
	\subfloat[Baseline]{\includegraphics[scale=0.11]{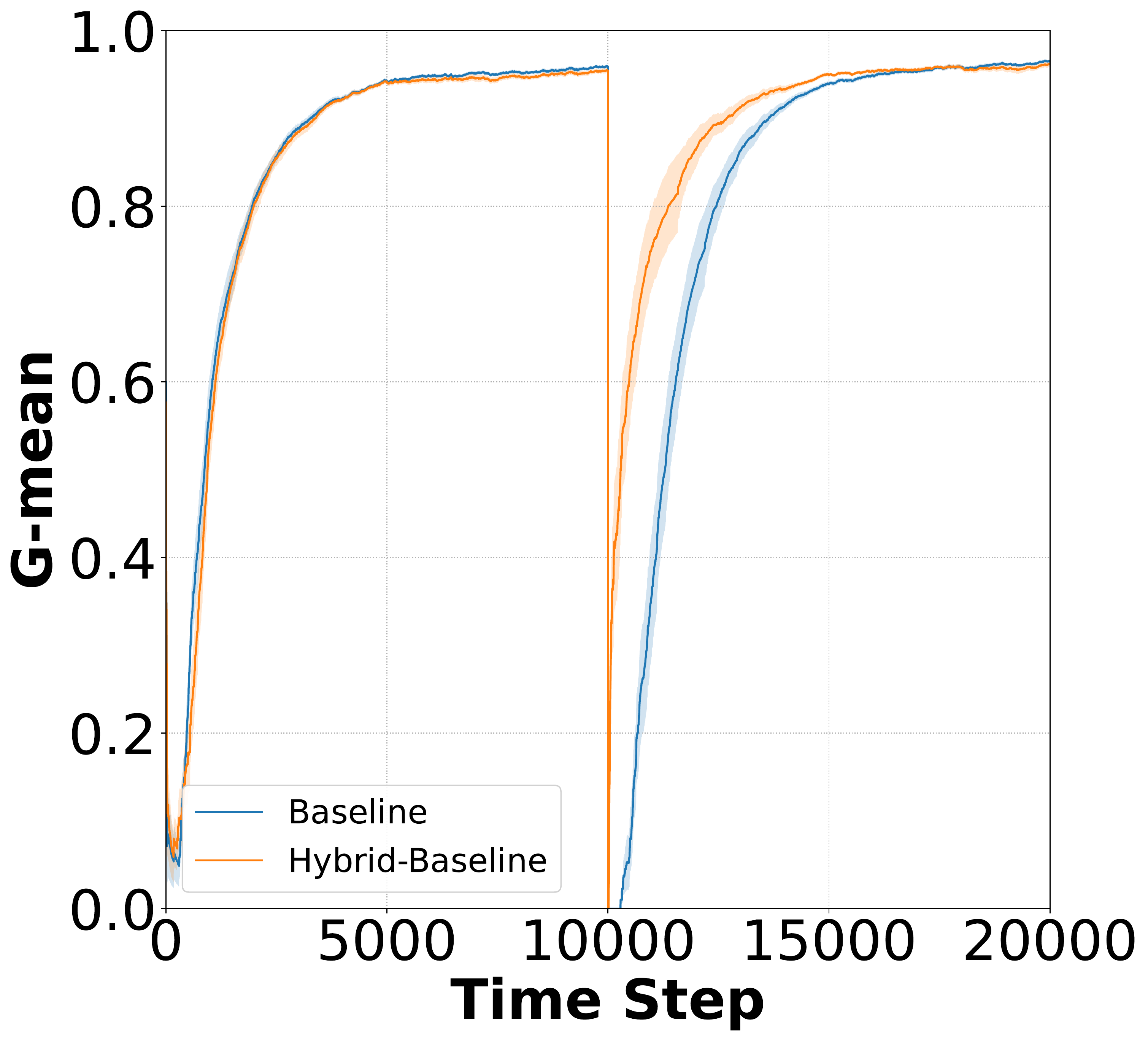}%
		\label{fig:sea01_drift_baseline}}
	\subfloat[Sliding]{\includegraphics[scale=0.11]{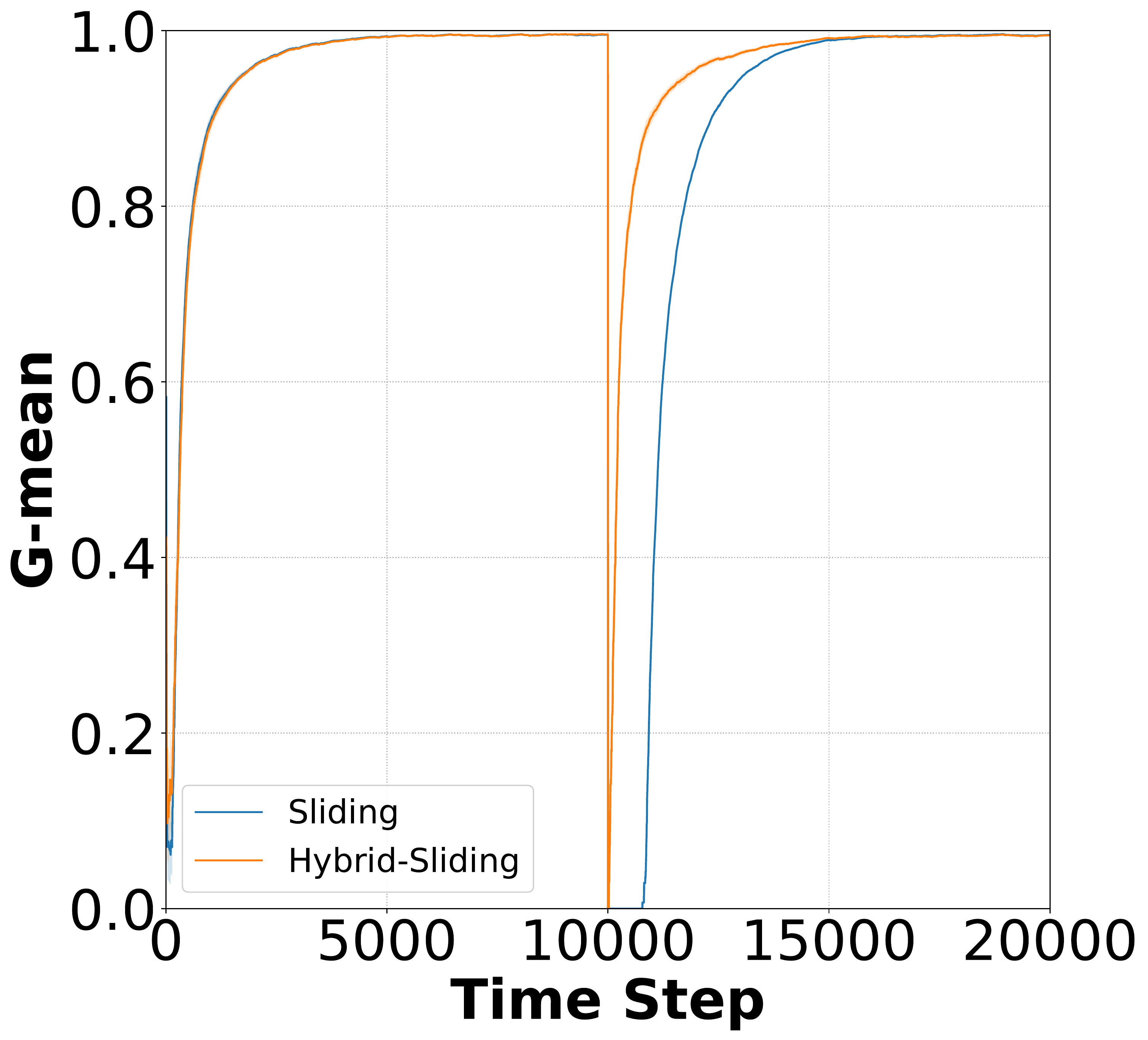}%
		\label{fig:sea01_drift_sliding}}
	\subfloat[Adaptive\_CS]{\includegraphics[scale=0.11]{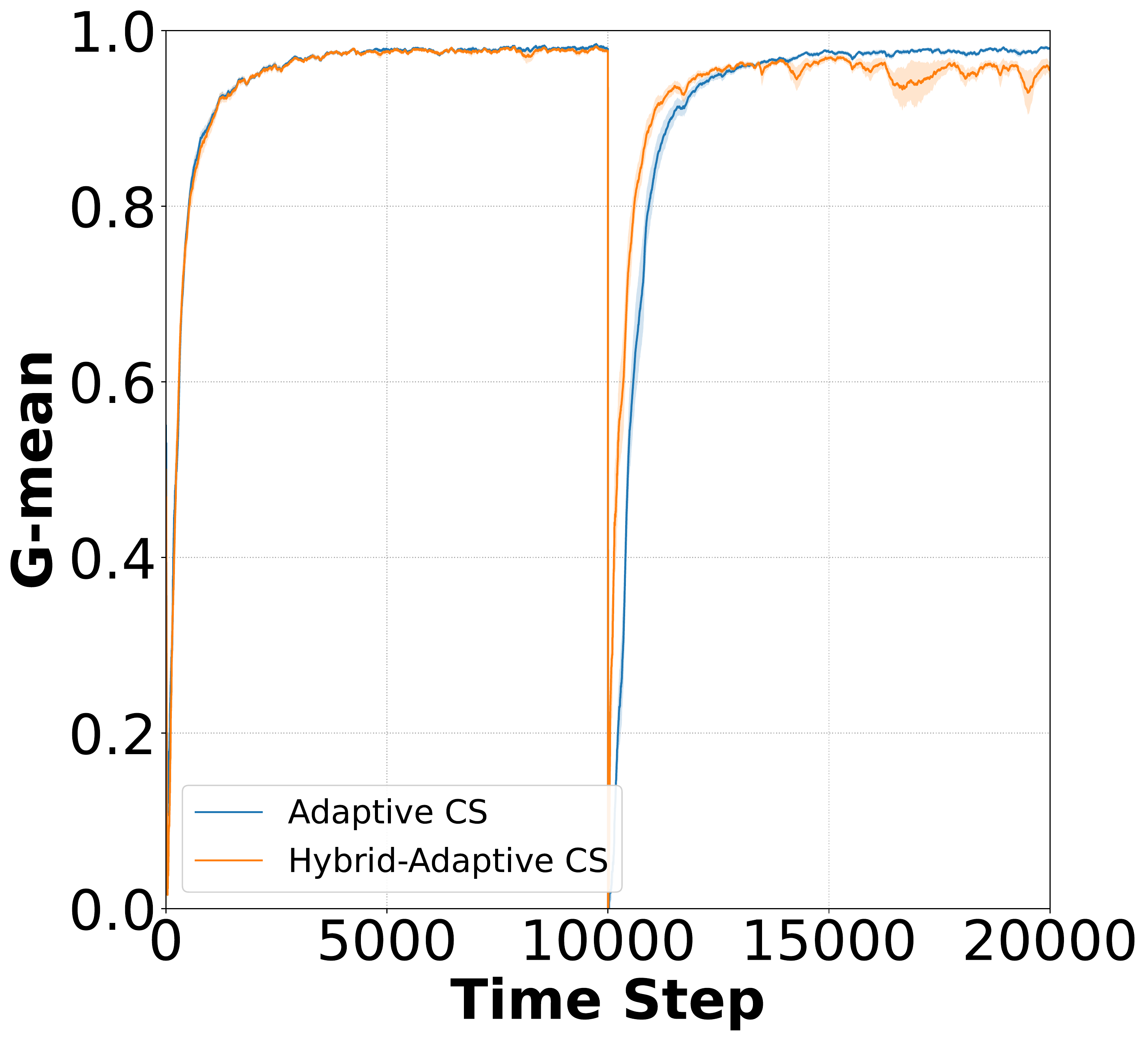}%
		\label{fig:sea01_drift_adaptive_cs}}
	\subfloat[OOB]{\includegraphics[scale=0.11]{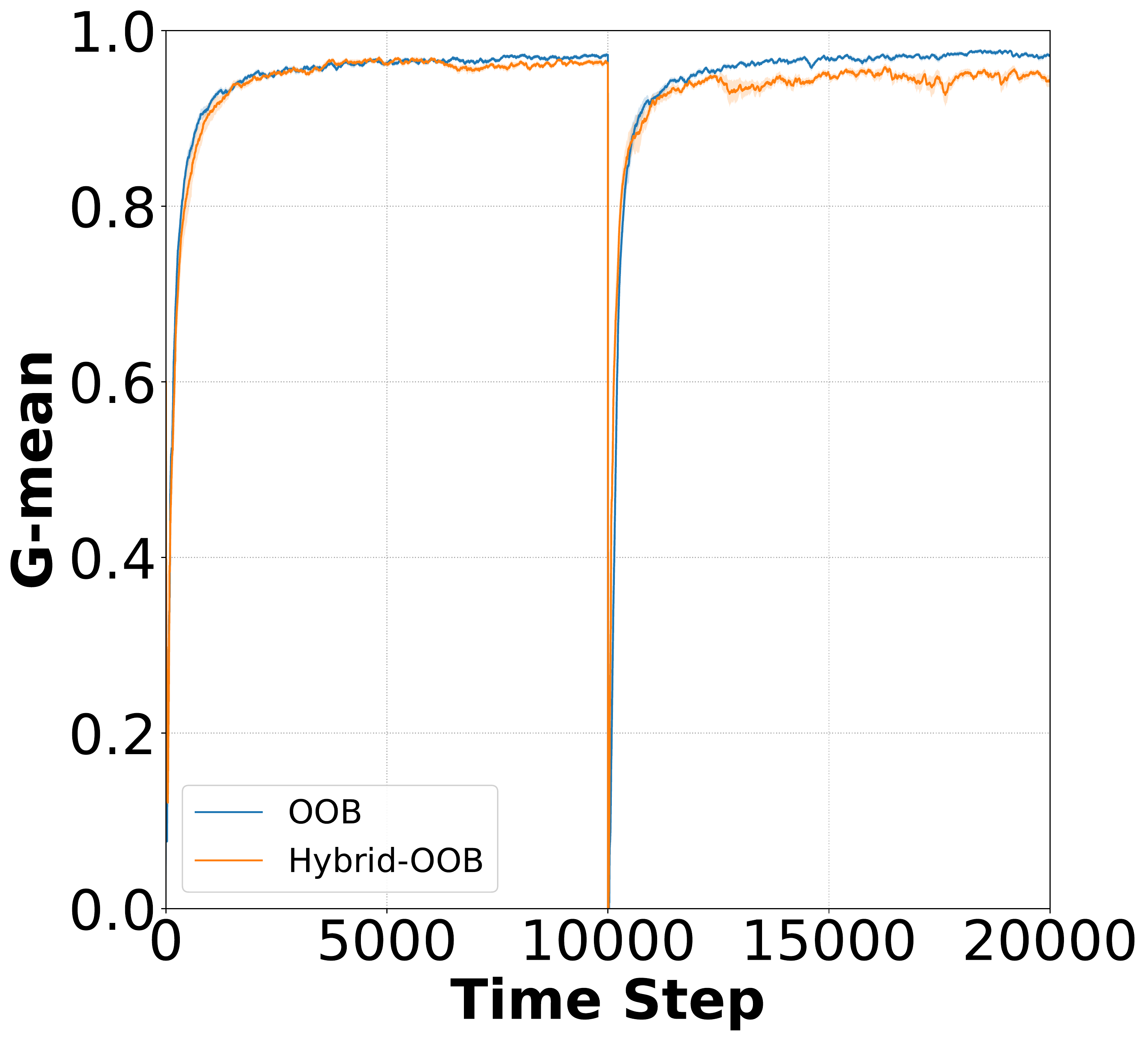}%
		\label{fig:sea01_drift_oob}}
	\subfloat[AREBA]{\includegraphics[scale=0.11]{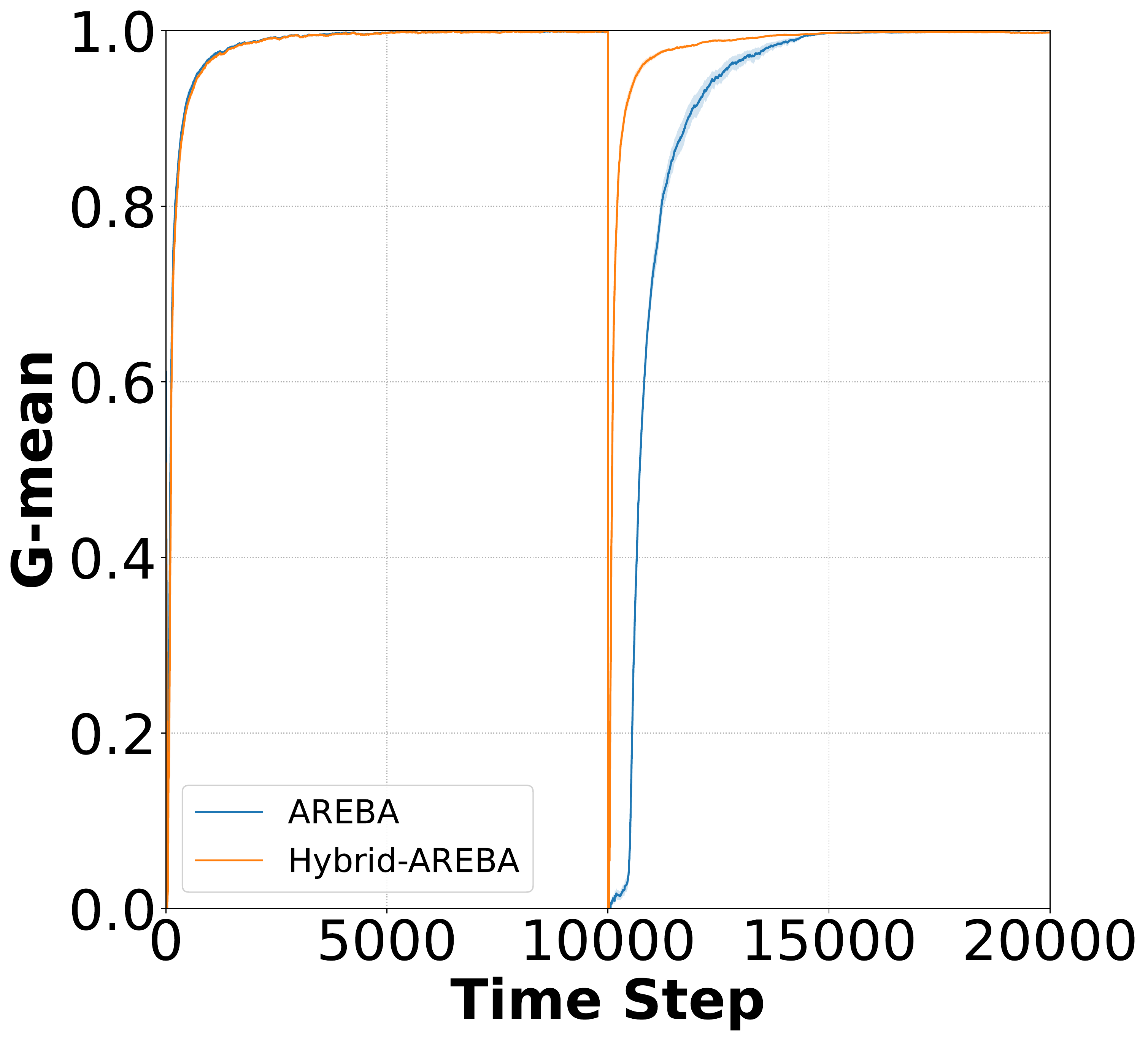}%
		\label{fig:sea01_drift_areba.}}
	
	\caption{The role of hybrid learning in the Sea dataset}\label{fig:role_sea}
\end{figure*}

\subsection{Datasets}\label{sec:datasets}
We consider three popular synthetic datasets, as they provide us the flexibility to control the drift characteristics (e.g., type and speed) and imbalance level, which are described shortly. These datasets represent benchmarks in the related literature.

\textbf{Circle} \cite{gama2004learning}: It has two features $x_1, x_2 \in [0, 1]$. The decision boundary is a circle with centre $(x_{1c}, x_{2c}) = (0.4, 0.5)$ and radius $r_c = 0.2$. Examples that fall inside the circle are classified as positive ($y=1$) and outside as negative ($y=0$). Fig.~\ref{fig:circle} depicts the dataset.

\textbf{Sine} \cite{gama2004learning}: It has two features in the range of $x_1 \in [0,2\pi]$ and $x_2 \in [-1,1]$. The decision boundary is $sin(x_1)$. Instances below the curve are classified as positive ($y=1$) and above the curve as negative ($y=0$). Rescaling has been performed so that both $x_1$ and $x_2$ are in $[0, 1]$. Fig.~\ref{fig:sine} depicts the dataset.

\textbf{Sea} \cite{street2001streaming}: It has two features $x_1, x_2 \in [0, 10]$. Instances that satisfy $x1 + x2 \leq 7$ are classified as positive, otherwise as negative. Rescaling has been performed so that $x1, x2 \in [0, 1]$. Fig.~\ref{fig:sea} depicts the dataset.

In all datasets, we consider drift that alters abruptly the posterior probability $p(y | x)$ of the data distribution. Specifically, drift causes a ``concept swap''. Furthermore, we consider two imbalance levels, mild ($10\%$) and severe $1\%$.

To demonstrate the robustness and general applicability of the Active part of the proposed hybrid approach, the same hyper-parameters are used in all datasets and for all compared methods. These are,  $waiting\_time = 500, expire\_time = 100, W_{scores} = 500$, $W_{dd} = 100$, $\beta_{drift} = 5.0$, $\beta_{warn} = 3.0$, and $\beta_{improve} = 2.5$.

\subsection{Class imbalance methods}
Our study is diverse as we consider methods that don't adopt any mechanisms to address imbalance (Baseline, Sliding), and methods that adopt such mechanisms, e.g., cost-sensitive learning (Adaptive\_CS) and resampling (OOB, AREBA).
 
\begin{table}[b!]
	\centering
	\resizebox{\columnwidth}{!}{%
		\begin{tabular}{lc}
			\hline
			\textbf{} & \textbf{Value} \\ \hline
			\textbf{Number of hidden layers} & 1 \\
			\textbf{Number of units} & {8} \\
			\textbf{Optimiser} & Adam \cite{kingma2014adam}\\
			\textbf{Learning rate} & 0.01 \\
			\textbf{Number of epochs} & 50 (Active) / 1 (Passive, except OOB) \\
			\textbf{Hidden activation} & Leaky ReLU \cite{maas2013rectifier}\\
			\textbf{Output activation} & Sigmoid \\
			\textbf{Loss function} & Binary cross-entropy \\
			\textbf{Mini-batch size} & 1 (one-pass learners) / memory size\\
			\textbf{Weight initialiser} & He Normal \cite{he2015delving}\\ \hline
		\end{tabular}%
	}
	\caption{Hyper-parameters of the base NN}
	\label{tab:nn_hyperparameters}
\end{table}

 For a fair comparison, we apply the same explicit drift detection method (Active part in Algorithm~\ref{alg:hybrid-areba}) to all methods. Also, the same base classifier is used by all methods, which is a standard fully-connected neural network as shown in Table~\ref{tab:nn_hyperparameters}. Notice that for training in the Active part (Line 32) we use 50 epochs, while for incremental training in the Passive part (Line 45) we use 1 epoch to avoid potential overfitting.

\textbf{Hybrid-Baseline.} A baseline method that performs incremental learning, with no explicit mechanism to handle imbalance. It is a one-pass learner as it doesn't use any memory.

\textbf{Hybrid-Sliding.} It uses a sliding window of size $W=1000$ to implicitly address drift, but no mechanism to address imbalance. It is not a one-pass learner (due to memory).

\textbf{Hybrid-Adaptive\_CS.} A state-of-the-art method that uses CSOGD and CID (Section~\ref{sec:related_imbalance}). The initial values are  $\gamma_p = 0.95$ and $\gamma_n = 0.05$, as suggested by its authors. To deal with stability issues, the ratio is bounded within $c \in [1,50]$ and is updated every $250$ steps. It is a one-pass incremental learner.

\textbf{Hybrid-OOB.} A state-of-the-art resampling method described in Section~\ref{sec:related_imbalance}. It also uses the CID method. It is a one-pass incremental learner. For computational reasons, we don't use its ensemble version.

\textbf{Hybrid-AREBA.} The proposed method described in Section~\ref{sec:method} and shown in Algorithm~\ref{alg:hybrid-areba}.

\subsection{Performance metrics and Evaluation method}
We use the geometric mean (G-mean) performance metric, which is insensitive to class imbalance \cite{he2008learning}. G-mean evaluates the degree of inductive bias in terms of a ratio of positive accuracy (recall) and negative accuracy (specificity). It has some desirable properties as it is high when both recall and specificity are high, and when their difference is small.

For evaluation we adopt the popular \textit{prequential error with fading factors} method, which is widely used by the community. The method has been proven to converge to the Bayes error for learning algorithms under stationary conditions \cite{gama2013evaluating}. It does not require a holdout set and the learning algorithm is always tested on unseen data. The fading factor is set to $\theta = 0.99$. We calculate the prequential G-mean in every step averaged over 20 repetitions, and we plot the error bars displaying the standard error around the mean.

\section{Experimental Results}\label{sec:exp_results}

\subsection{Role of hybrid learning}
We examine here the role of hybrid learning in all datasets for all compared methods. Fig.~\ref{fig:role_circle} compares the performance of Baseline, Sliding, Adaptive\_CS, OOB and AREBA with hybrid learning and without (i.e., incremental learning only). Similarly, Figs.~\ref{fig:role_sine} and \ref{fig:role_sea} show the performance in the Sine and Sea datasets. We can extract the following conclusions.
\begin{itemize}
	\item The effect of hybrid learning is prominent, and in almost all cases hybrid learning significantly outperforms the use of incremental learning on its own.
	
	\item The improvement of hybrid learning might differ for each method. For example, in the Sine dataset (Fig.~\ref{fig:role_sine}), while the improvement is large for Baseline, Adaptive\_CS and AFREBA, it is much smaller for OOB.
	
	\item In some cases, such as, in Figs.~\ref{fig:sine01_drift_sliding}, \ref{fig:sea01_drift_adaptive_cs} and \ref{fig:sea01_drift_oob}, hybrid learning causes a small decline and / or fluctuations in the performance. This is attributed to the false alarms raised by the concept drift detection method. Recall from Section~\ref{sec:datasets} that we have kept the same parameters for the drift detection method in all settings to demonstrate its practicality and general applicability ($\beta_{drift} = 5.0$, $\beta_{warn} = 3.0$, $\beta_{improve} = 2.5$). Had hyper-parameter tuning been performed for every dataset-method case, we would observe fewer false alarms.
\end{itemize}

 \begin{figure}[t]
	\centering
	
	\subfloat[Circle 10\%]{\includegraphics[scale=0.14]{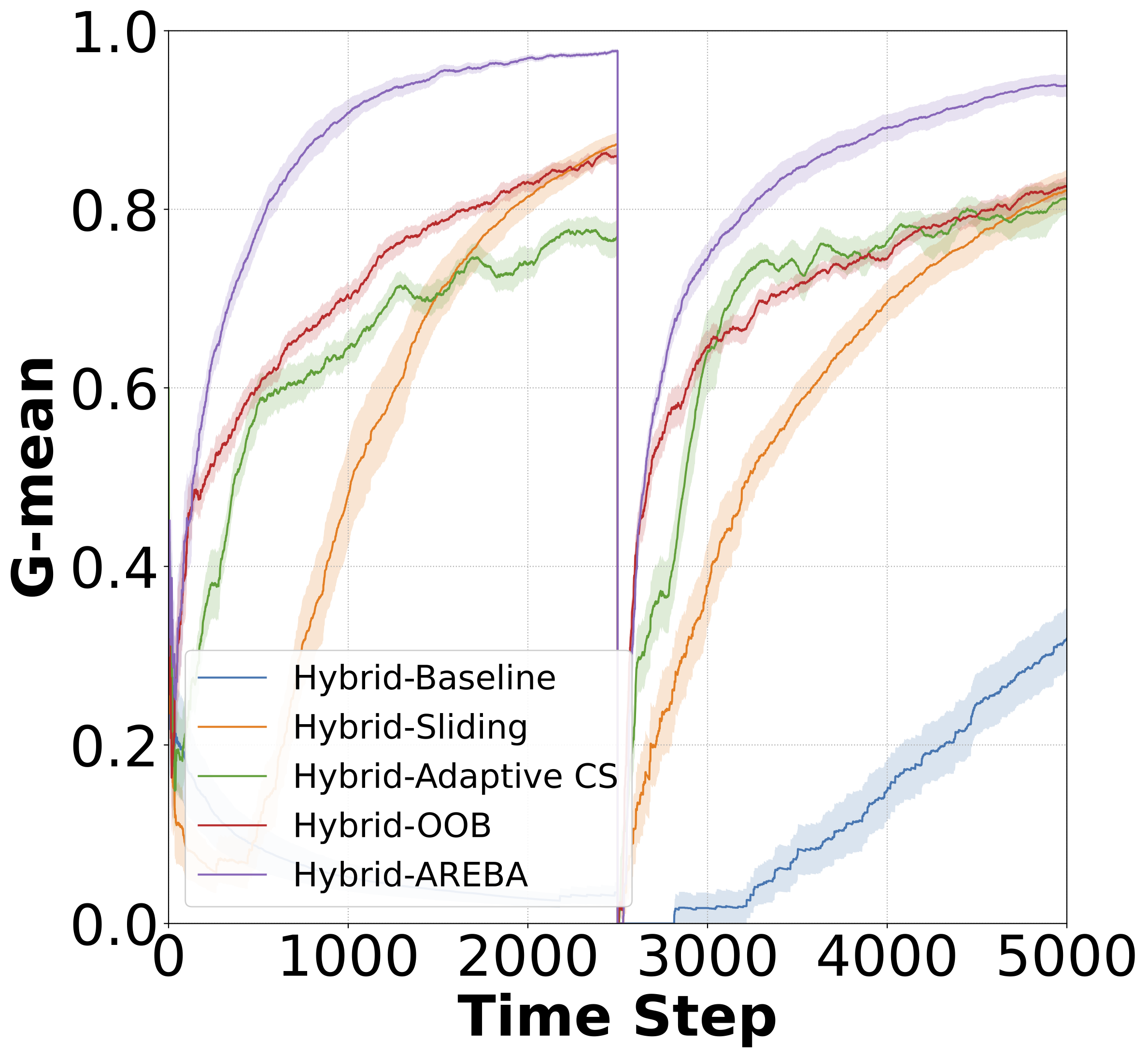}%
		\label{fig:circle01}}
	\subfloat[Circle 1\%]{\includegraphics[scale=0.14]{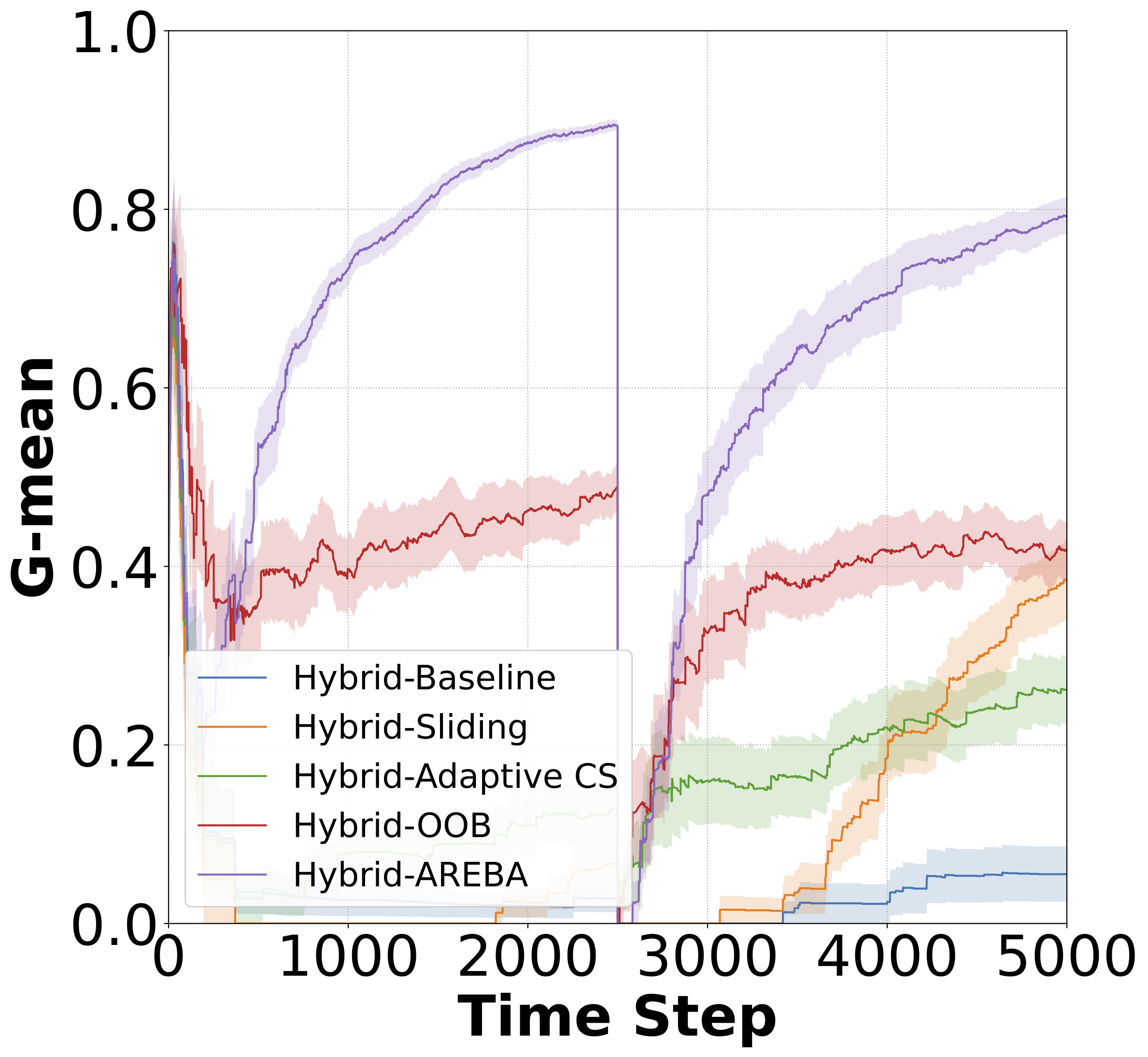}%
		\label{fig:circle001}}

	\subfloat[Sine 10\%]{\includegraphics[scale=0.14]{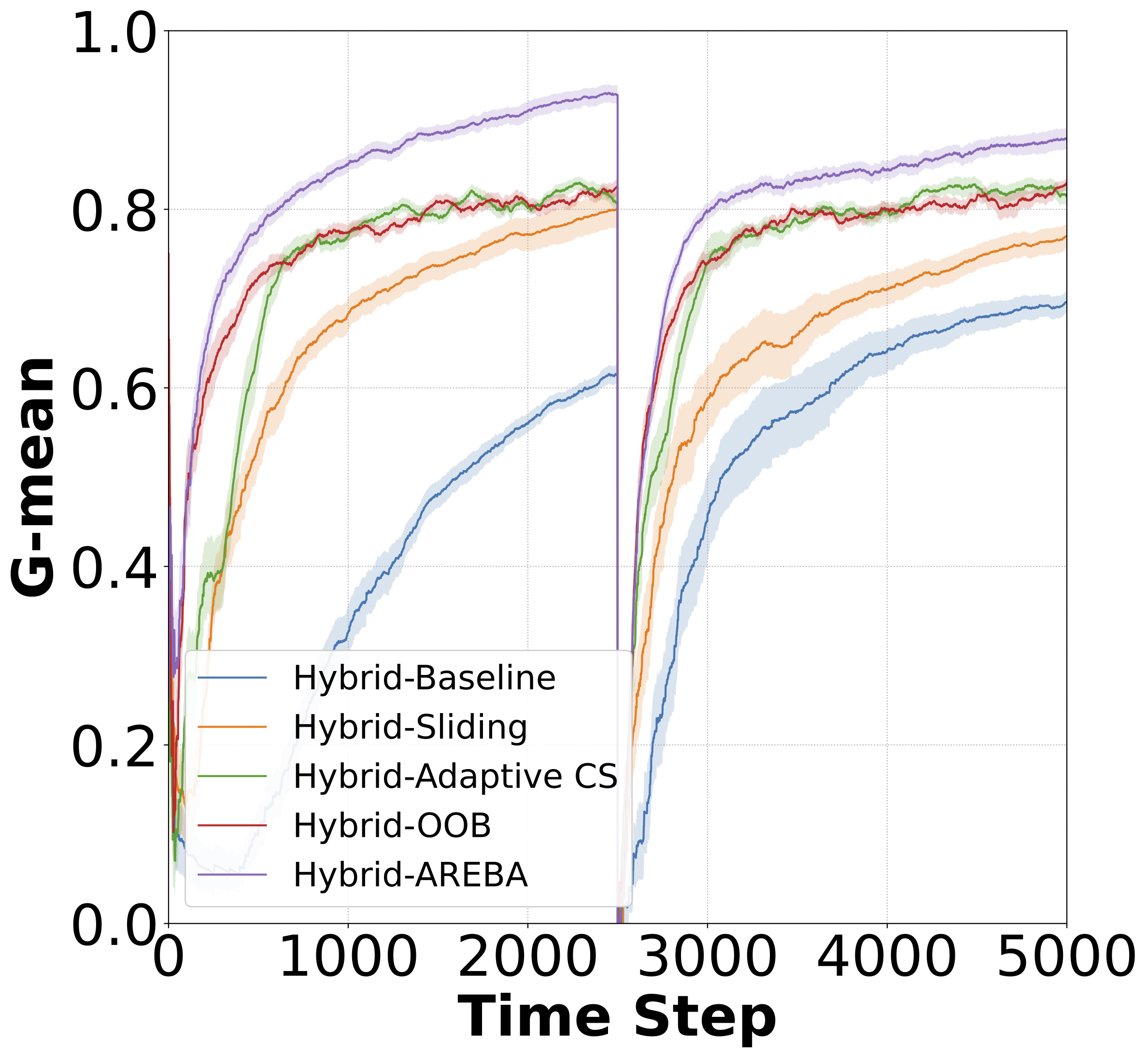}%
		\label{fig:sine01}}
\subfloat[Sine 1\%]{\includegraphics[scale=0.14]{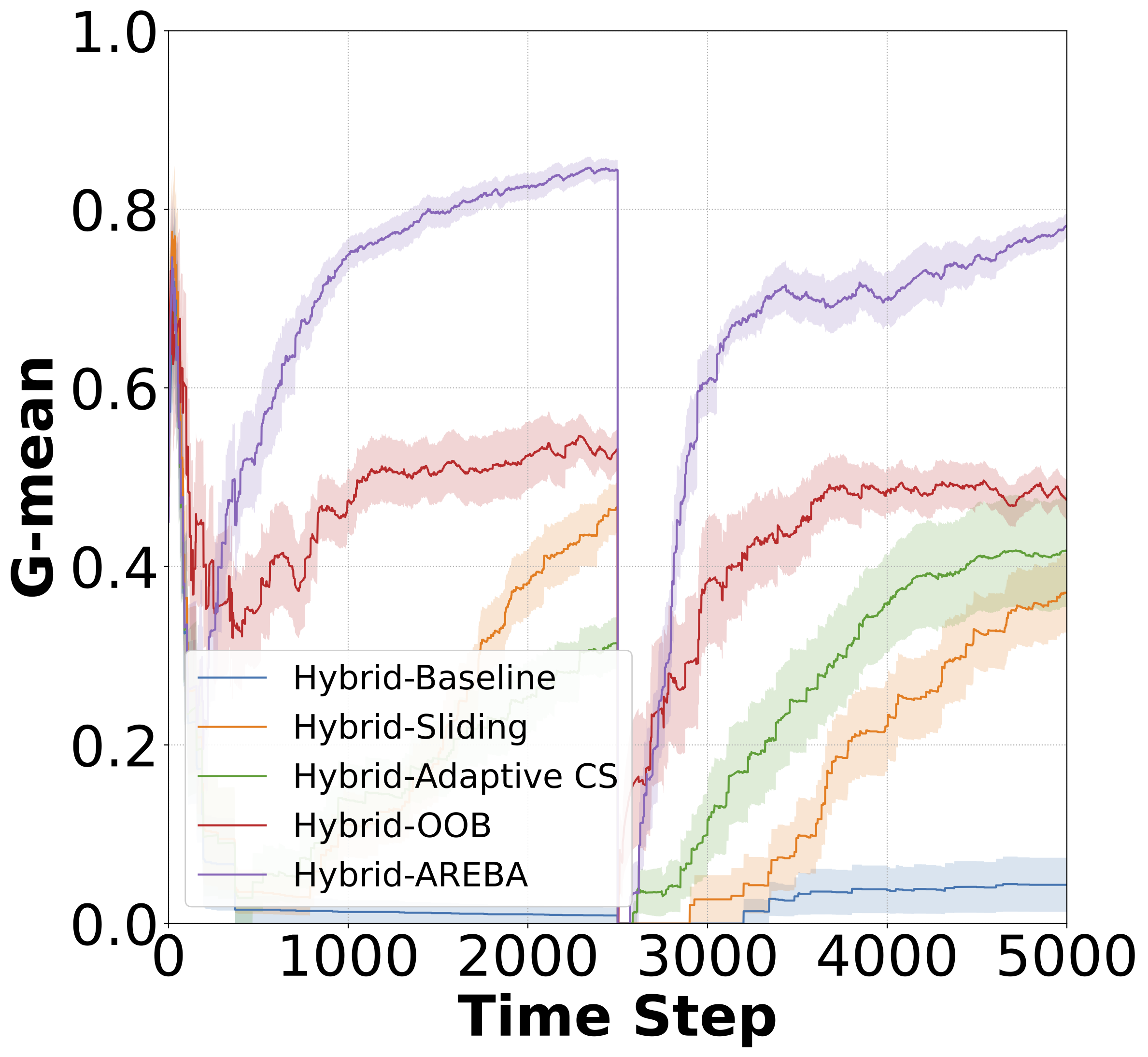}%
		\label{fig:sine001}}

	\subfloat[Sea 10\%]{\includegraphics[scale=0.14]{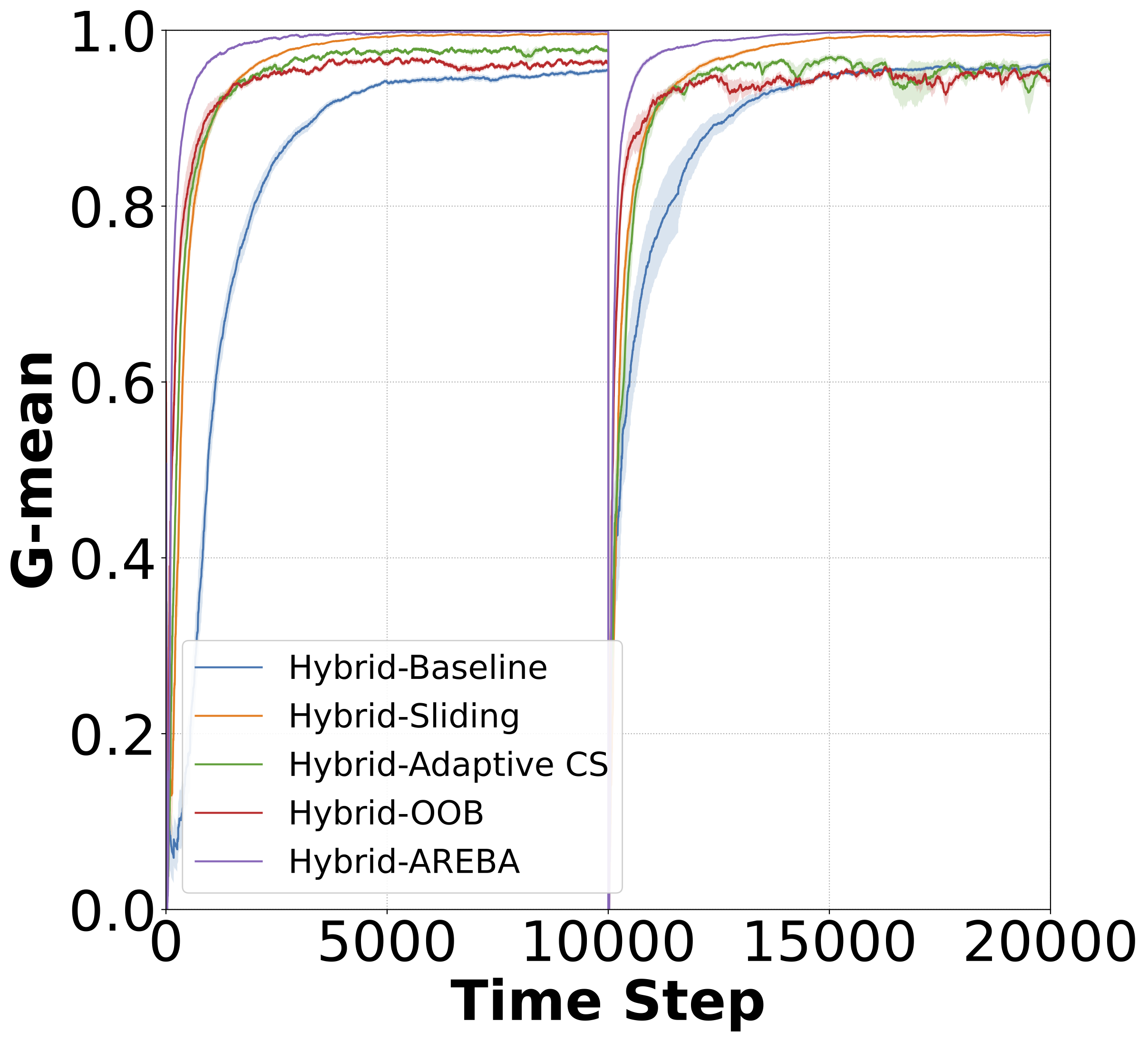}%
		\label{fig:sea01}}
	\subfloat[Sea 1\%]{\includegraphics[scale=0.14]{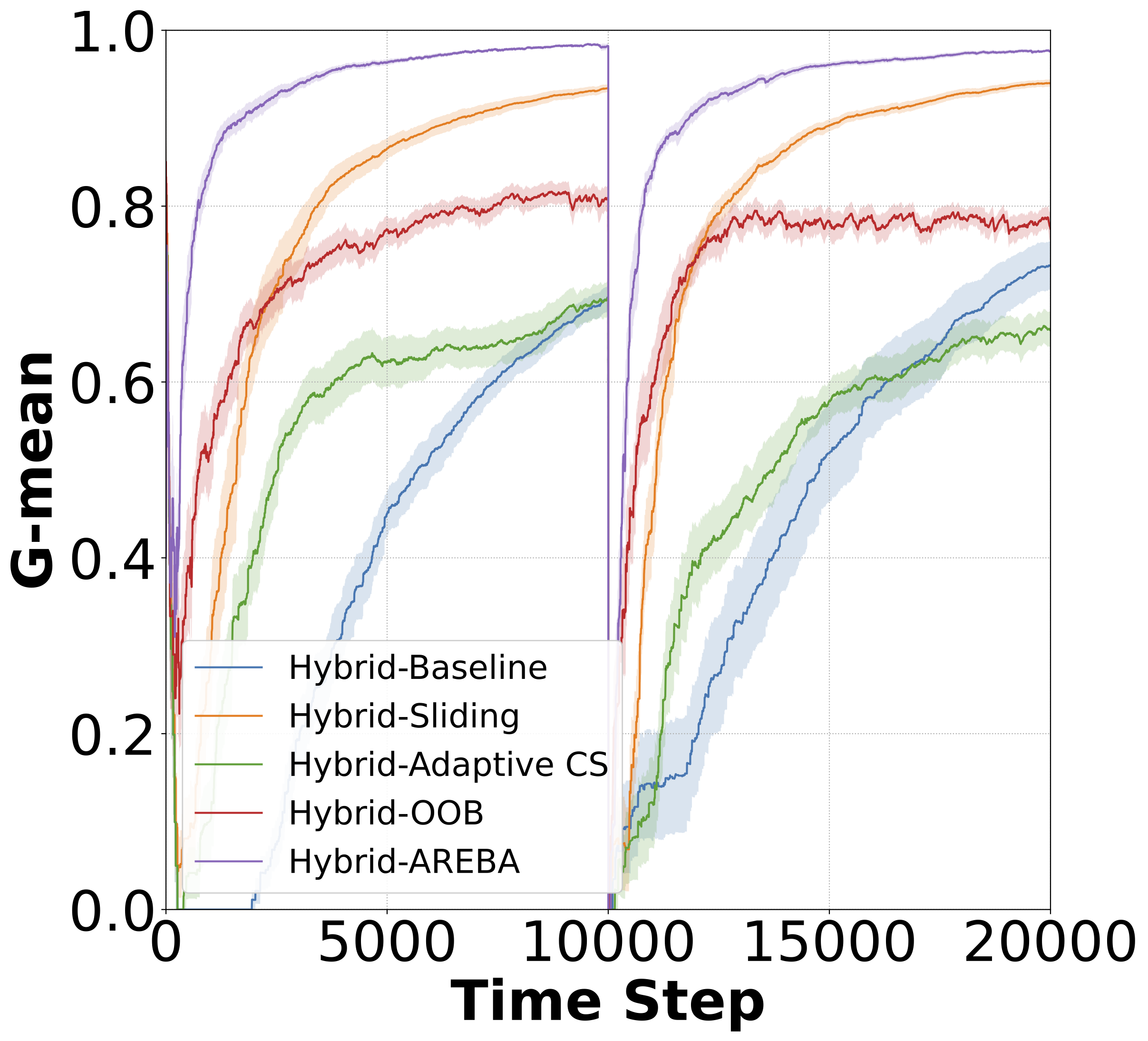}%
		\label{fig:sea001}}
	
	\caption{Comparative study}\label{fig:comparison}
\end{figure}

\subsection{Comparative study}
In this section we perform a comparative study of the different methods that address class imbalance. Fig.~\ref{fig:comparison} shows the performance for the hybrid versions of Baseline, Sliding, Adaptive\_CS, OOB, and AREBA. The first, second, and third rows correspond to Circle, Sine, and Sea datasets respectively. The first column corresponds to 10\% class imbalance, while the second to the more challenging case of 1\% imbalance.

We can extract the following conclusions.
\begin{itemize}
	\item In all datasets, under both imbalance levels, and either prior or after the drift, the proposed Hybrid-AREBA significantly outperforms the rest. Importantly, even after providing more time, no other method can equalise its performance in Fig.~\ref{fig:sea001}, and only Sliding equalises its performance in Fig.~\ref{fig:sea01}. This is attributed to the following reasons. Apart from the explicit drift detection method, Hybrid-AREBA performs resampling to handle imbalance, and it is memory-based, i.e., it has an implicit way to handle drift during incremental learning.
		
	\item Methods that have a mechanism to address imbalance (OOB, Adaptive\_CS), typically, yield the second best performance. When the imbalance is 10\% (first column), their performance is comparable. When the imbalance is 1\%, OOB significantly outperforms Adaptive\_CS.
	
	\item Sliding performs the next best performance, attributed to the fact that it doesn't have a mechanism to handle imbalance. Notable exceptions are observed in Figs.~\ref{fig:sea01} and \ref{fig:sea001}. This is attributed to the fact that drift is the more important problem in this case rather than imbalance.
	
	\item Expectedly, Baseline yields the worst performance.
\end{itemize}

\section{Conclusion}\label{sec:conclusion}
Extracting patterns from data streams poses major challenges, including, concept drift (data nonstationarity) and class imbalance (data skewness). Contrary to the traditional avenue to address drift, we propose the hybrid active-passive method HAREBA, which effectively addresses concept drift, is robust to severe imbalance, and significantly outperforms strong baselines and state-of-the-art methods. Future work will examine more advanced drift detection mechanisms, such as, using statistical tests, and evaluate the proposed method in real-world datasets under various drift characteristics.

\bibliographystyle{IEEEtran}
\bibliography{mybib}

\end{document}